\documentclass{article}
\usepackage[margin=0.85in]{geometry}
\usepackage{graphicx} % Required for inserting images
\usepackage{booktabs}
\usepackage{enumitem}
\usepackage[table]{xcolor}
\usepackage{array}
\usepackage{natbib}
\usepackage{rotating}
\usepackage{lscape}
\usepackage{amsmath}
\usepackage{amsfonts}
\usepackage{amssymb}
\usepackage{changepage}
\usepackage{pdflscape}
\usepackage{algorithm}
\usepackage{algpseudocode}
\usepackage{listings}
\usepackage{comment}
\usepackage{xcolor}

\definecolor{codegreen}{rgb}{0,0.6,0}
\definecolor{codegray}{rgb}{0.5,0.5,0.5}
\definecolor{codepurple}{rgb}{0.58,0,0.82}
\definecolor{backcolour}{rgb}{0.95,0.95,0.92}

\usepackage{longtable}
\usepackage{xcolor}
\usepackage{setspace}

\usepackage{hyperref}
\hypersetup{
    colorlinks=true,
    linkcolor=black,
    filecolor=black,
    urlcolor=black,
    citecolor=black
}

\title{Prompt Stability Scoring for Text Annotation with Large Language Models}
\author{Christopher Barrie,$^{1}$ Elli Palaiologou,$^{2}$ Petter Törnberg$^{3}$\\[0.5em]
\normalsize{$^{1}$Department of Sociology, New York University}\\
\normalsize{$^{2}$Independent Researcher}\\
\normalsize{$^{3}$Institute for Logic, Language, and Computation, University of Amsterdam}}
\date{\today}

\begin{document}

\maketitle

\begin{abstract}
\noindent Researchers are increasingly using language models (LMs) for text annotation, typically relying on prompts that instruct the model to return a given output. LM outputs can nonetheless be vulnerable to minor prompt variations, raising concerns about the reliability and reproducibility of classification routines. Stability is a necessary but not sufficient condition for both accuracy and reproducibility. To diagnose this, we propose the Prompt Stability Score (PSS), an automated procedure for measuring agreement across repeated runs and semantically similar prompt variants, and provide an accompanying Python package, \texttt{promptstability}. Using six datasets and twelve outcomes, we generate $\sim$3.1m annotations ($\sim$300m input tokens) to characterize when prompt-based pipelines are stable versus fragile. We conclude with practical recommendations for applied research.

\end{abstract}
\clearpage
\doublespacing
\section{Introduction}

Given recent advances in natural language processing, scholars have begun debating whether human coders may be replaced by machine-learning tools \citep{gilardi2023chatgpt, grossmann2023}. Specifically, commentary has focused on large language models (LMs), which rely on minimal researcher input and model training for the classification of different types of data \citep{ziems2024can, tornberg2023use}. Given the power of many LMs, researchers have proposed that so-called zero-shot approaches are capable of overcoming even demanding classification tasks \citep{le2023uncovering, gilardi2023chatgpt}. Importantly, these approaches represent a departure from more conventional training and testing paradigms in supervised learning: we can now accurately classify data just by \textit{telling a model what we want it to do}.

Zero- or few-shot designs require just a ``prompt" to guide the model in its classification decisions. This prompt often takes the form of a short phrase or set of phrases indicating the desired output \citep{tornberg2023chatgpt}. Using these techniques, a large number of contributions have now demonstrated that LMs are capable of diverse classification or annotation tasks \citep{gilardi2023chatgpt, tornberg2023chatgpt, bang2023multitask, qin2023chatgpt,goyal2022news, chiang2023can, grossmann2023, ziems2024can}.

Absent from much of the recent applied literature is a routine, reliability-style check on LM-based annotation pipelines \citep{reiss2023testing, Ollion2023, sclar2023quantifying}. This omission matters because success with a single prompt does not imply that small, substantively equivalent rephrasings will yield the same labels. If outputs shift across semantically similar prompts, the procedure is fragile and difficult to reproduce. Our method provides an automated stability diagnostic that can be run \emph{before} any accuracy validation. High stability does not guarantee correctness---it simply indicates that the pipeline is internally consistent under minor prompt perturbations. Low stability, by contrast, is a warning that the prompt, construct definition, data, and/or model settings should be revisited before committing to costly validation steps such as manual coding, as we go on to detail below.

We introduce a general framework---and an accompanying Python package, \texttt{promptstability}---that automates stability diagnostics for LM-based text annotation. Our focus is \textit{prompt stability}: the extent to which a model assigns the same labels when (i) the \emph{same} prompt is run repeatedly and (ii) \emph{semantically equivalent} prompt variants are used. Starting from a user-supplied baseline prompt, our algorithm generates a controlled set of prompt paraphrases and applies each prompt (and repeated runs of a fixed prompt) to the same items. We then quantify agreement using an established reliability coefficient, treating repeated runs or prompt variants as partial analogues of human inter-rater reliability scores. The resulting inter- and intra-Prompt Stability Scores (PSS) provide a practical diagnostic of whether an annotation pipeline is robust to minor prompt perturbations---and therefore whether it is likely to be reproducible under realistic variations in prompt wording or model behavior.

\section{Why prompt instability occurs}

Large language models (LMs) generate outputs one token at a time. At each step, the model assigns probabilities to candidate next tokens conditional on the prompt and any text generated so far. Prompt-based ``classification'' is therefore implemented through the same next-token generation mechanism as open-ended text generation: even when we request a constrained label (e.g., \texttt{0/1}, a category name, or a numeric score), the model arrives at that label through token-level prediction rather than by retrieving a fixed class from an internal lookup table.

This has an immediate implication for reproducibility. Because the model's output distribution is conditioned on the prompt, seemingly minor rephrasings can change which cues are emphasized (e.g., what counts as evidence, how strict criteria should be, or how edge cases are handled). When an item is close to a decision boundary---that is, when multiple labels remain plausible given the text and instructions---small shifts in token probabilities can flip the most likely completion. The effect is especially salient for discrete outputs, where small probability changes can produce different final labels \citep{sclar2023quantifying}.

A second, independent source of variation is \emph{decoding stochasticity} in the annotation model. Many inference pipelines sample from the model's token distribution rather than deterministically selecting the single most probable token at each step. This sampling behavior is governed by decoding parameters such as \emph{temperature} and \emph{top-$p$}: lower values yield more conservative and repeatable outputs, while higher values increase randomness and diversity. As a result, repeated runs of the same prompt (even with these decoding parameters fixed at low values) on the same input can yield different annotations. This run-to-run variability is what we later quantify as \emph{intra}-prompt stability.

A third, less visible source of randomness arises from nondeterminism in the model’s forward pass itself. Modern LM inference relies on highly parallelized GPU execution, mixed-precision arithmetic, speculative decoding, and, in some architectures, stochastic expert routing, all of which can introduce small perturbations in intermediate activations. Because these perturbations are propagated through many nonlinear layers, they can lead to substantively different logit distributions across runs, particularly for borderline cases. Consequently, even nominally identical inference settings may produce different output distributions prior to decoding, further contributing to instability in LM-based annotation.

What we go on to describe as \emph{inter}-prompt stability analysis, by contrast, relies on varying the \emph{wording of the prompt itself} while holding the annotation model fixed. We generate semantically similar prompt variants using an external paraphrasing model whose \emph{temperature} controls how conservative versus diverse those variants are. We describe this prompt-generation procedure, and the distinction between annotation-model temperature and paraphraser temperature, in the Methods section below.

Taken together, these sources of variation are what motivate the diagnostic we propose in this paper. If a model returns different labels when (i) the \emph{same} prompt is repeated across runs, or (ii) \emph{semantically equivalent} prompt variants are used, while holding the annotation model and decoding settings fixed, then the classification pipeline is sensitive to minor perturbations in a way that undermines reproducibility. The next section links this idea to classic inter- and intra-coder reliability and shows how we operationalize prompt stability using standard reliability statistics.

\section{Coder reliability and prompt stability}

Prompt stability is naturally framed using the logic of inter- and intra-coder reliability. In traditional content analysis, \emph{intra}-coder reliability captures whether the \emph{same} coder assigns the same label when coding the same items multiple times, while \emph{inter}-coder reliability captures whether \emph{different} coders agree when coding the same items. These provide a check on replicability by showing that results are not unduly sensitive to who codes the data or when the coding occurs \citep{o2020intercoder, belur2021interrater, lamprianou2023measuring}.

We adapt this logic to prompt-based annotation with LMs. For \textbf{intra}-prompt stability (intra-PSS), we run the \emph{same} prompt repeatedly on the \emph{same} items and treat each run as a ``rater.''\footnote{With the caveat that LM variability need not mirror human variability; see \cite{barrie2024replication}.} For \textbf{inter}-prompt stability (inter-PSS), we apply multiple \emph{semantically equivalent} prompt variants to the same items and treat each prompt variant as a ``rater.'' In both cases, we summarize agreement using Krippendorff's $\alpha$, but the rater-equivalents are induced by design (runs or prompt variants), not by recruiting additional human coders. The contribution of this paper is therefore not a new agreement statistic, but an automated framework for \emph{eliciting} and \emph{summarizing} these stability diagnostics under controlled prompt perturbations.

We focus on stability rather than accuracy because stability is a precondition for reproducibility. A highly stable classifier can still be wrong; however, if outputs shift when the \emph{same} prompt is re-run or when trivially different (but equivalent) prompts are used, then any downstream estimates inherit a fragility that undermines replication. PSS is therefore intended as an early-stage diagnostic: it helps researchers determine whether a prompt-based annotation pipeline is robust enough to warrant subsequent validation against ground truth.

Low stability can arise for reasons that overlap with classic sources of low human reliability, but the mapping is imperfect for LMs. Table~\ref{table:lowicr_ps} summarizes a practical taxonomy: instability may reflect ambiguity in criteria (prompt wording), difficulty or noise in the data, under-specified or contested constructs, and/or model capability.

\begin{table}[ht]
\centering
\renewcommand{\arraystretch}{1.3}
\begin{tabular}{p{2.5cm}p{4.25cm}p{4.25cm}}
\toprule
\textbf{Aspect} & \textbf{Coder Reliability} & \textbf{Prompt Stability} \\
\midrule
\begin{tabular}[t]{@{}l@{}}(1) Criteria \\ Ambiguity\end{tabular}
    & Ambiguous coding instructions lead to inconsistent interpretations.
    & Ambiguous prompt design can lead to inconsistent language model outputs. \\
\addlinespace[0.5em]

\begin{tabular}[t]{@{}l@{}}(2) Data \\ Complexity\end{tabular}
    & Complex, messy, or inappropriate data can lead to unreliable classifications by coders.
    & Complex, messy, or inappropriate data may result in unreliable classifications by the language model. \\
\addlinespace[0.5em]

\begin{tabular}[t]{@{}l@{}}(3) Outcome \\ Complexity\end{tabular}
    & The construct may lack a stable definition for reliable human coding.
    & The construct may lack a stable definition for reliable language model classification. \\
\addlinespace[0.5em]

\begin{tabular}[t]{@{}l@{}}(4) Skill / \\ Capability\end{tabular}
    & Varying coder experience, skill, or training may lead to unreliable classifications.
    & Variability in language model capability may lead to unreliable classifications. \\
\bottomrule
\end{tabular}
\caption{Factors affecting low coder reliability and prompt stability in research contexts}
\label{table:lowicr_ps}
\end{table}

Unlike in human coding---where coder competence can sometimes be established and remaining disagreement attributed to unclear criteria, problematic data, or an ill-defined construct---``model capability'' is harder to assess. In what follows, we run targeted stress tests---varying prompt wording, the amount of information in the input, and the outcome definition---because these are realistic sources of variation in applied work and plausible ways instability could arise. We use the resulting changes in PSS to characterize how sensitive an annotation pipeline is to these perturbations.

\section{Current Approaches}

Prior work relevant to LM reliability spans: (i) prompt-variation and calibration methods that use sensitivity in outputs to probe uncertainty or improve performance \citep{chen_quantifying_2023, shin_autoprompt_2020, portillo_wightman_strength_2023, zhao_calibrate_2021, arora_ask_2022, chen_relation_2023, schick_exploiting_2021, tian_just_2023, feffer2024prompt}; (ii) prompt-engineering strategies (e.g., Chain-of-Thought and variants) that search over prompt designs to improve task performance \citep{wei2022chain, liu2021generated, long2023large, wang2022self, madaan2023self, perez2021true, liu2023pre}; and (iii) applied demonstrations of strong LM annotation performance alongside evidence that outputs can be brittle to prompt wording and formatting \citep{gilardi2023chatgpt, tornberg2023chatgpt, bang2023multitask, qin2023chatgpt, goyal2022news, chiang2023can, grossmann2023, ziems2024can, dentella2023systematic, bisbee2023artificially, sclar2023quantifying}. Existing mitigations often rely on prompt perturbation and aggregation across variants \citep{tornberg2023chatgpt, ziems2024can}.

What is missing is a general, automated diagnostic that quantifies prompt sensitivity as a reproducibility risk---a gap that has motivated calls for new validation practices in LM-assisted social science \citep{ziems2024can, bail2024can}. Our framework addresses this gap by treating repeated runs or semantically equivalent prompt variants as rater equivalents and summarizing agreement with standard reliability statistics. This perspective is analogous to sensitivity analysis in earlier text-as-data methods, where small analytic choices could materially affect results \citep{denny2018text}.

\section{Design}
\subsection{Data}

We use six different datasets to validate our approach and we specify two versions of the outcome for each of these datasets. These are described in Table \ref{tab:datasets}. The data were selected for: a) being relevant to social science research;  b) spanning diverse types of construct; c) being adaptable to multiple operationalizations of that construct. The ``Analysis name" column is how we go on to refer to the data-outcome pairs below when discussing results. The ``Relevant variation" column links each dataset and task to two sources of low reliability identified in Table \ref{table:lowicr_ps}: data complexity (factor 2) and outcome complexity (factor 3). We address criteria ambiguity (factor 1) by evaluating the stability of semantically similar prompts. Although our main analyses rely on a single LM, researchers interested in assessing skill or capacity differences (factor 4) can easily replicate the analysis using any model of their choice—our package is compatible with all major LMs.\footnote{Example usage can be found on the package page: \url{https://pypi.org/project/promptstability/}.}.

The first dataset comes from \cite{van2020twitter} where we filter for all content posted by United States Senators on Twitter (now X) in the two months prior to the 2020 US Election.\footnote{This is the same data used in \cite{tornberg2023chatgpt}}. The second is a set of UK party political manifestos taken from \cite{benoit2016crowd}. The third is a set of New York Times news articles about a diverse range of topics taken from \cite{young2012affective}. The fourth is a set of tweets expressing stances toward individuals or topics in US politics from \cite{ziems2024can}.\footnote{These originally derive from the SemEval-2016 dataset provided by \cite{mohammad2016semeval}.} The fifth is a set of open-ended survey responses from the British Election Study used in \cite{mellon2024ais}. And the sixth is a series of text profiles of individual voters taken from the American National Election Study used in \cite{argyle2023out}.

\begin{landscape}
\begin{table}[htbp]
    \centering
    \renewcommand{\arraystretch}{.5}
    \begin{tabular}{p{2cm}p{2cm}p{6cm}p{4cm}p{4cm}}
        \toprule
        \textbf{Dataset} & \textbf{Source} & \textbf{Outcomes} & \textbf{Analysis name} & \textbf{Relevant variation} \\
        \midrule
        US Senator Tweets & van Vliet et al. (2020) & \begin{itemize}[nosep,leftmargin=*]
            \item Binary: Republican/Democrat
            \item Binary: Populist/Not Populist
        \end{itemize} & \begin{itemize}[nosep,leftmargin=*]
            \item Tweets (Rep. Dem.)
            \item Tweets (Populism)
        \end{itemize} & \begin{itemize}[nosep,leftmargin=*]
            \item Outcome complexity (easier)
            \item Outcome complexity (harder)
        \end{itemize} \\
        \addlinespace[0.5em]
        NYT News Articles & Young \& Soroka (2012) & \begin{itemize}[nosep,leftmargin=*]
            \item Categorical: Positive/Negative/Neutral (All Articles)
            \item Categorical: Positive/Negative/Neutral (Short Articles)
        \end{itemize} & \begin{itemize}[nosep,leftmargin=*]
            \item News
            \item News (Short)
        \end{itemize} & \begin{itemize}[nosep,leftmargin=*]
            \item Data complexity (harder)
            \item Data complexity (easier)
        \end{itemize} \\
        \addlinespace[0.5em]
        UK Manifestos & Benoit et al. (2016) & \begin{itemize}[nosep,leftmargin=*]
            \item Binary: Left-wing/Right-wing
            \item Integer: 1-10 Ideology Scale
        \end{itemize} & \begin{itemize}[nosep,leftmargin=*]
            \item Manifestos
            \item Manifestos Multi
        \end{itemize} & \begin{itemize}[nosep,leftmargin=*]
            \item Outcome complexity (easier)
            \item Outcome complexity (harder)
        \end{itemize} \\
        \addlinespace[0.5em]
        US Politics Tweets & Ziems et al. (2024) & \begin{itemize}[nosep,leftmargin=*]
            \item Binary: Stance For/Against Trump
            \item Categorical: For Trump/Against Trump/None
        \end{itemize} & \begin{itemize}[nosep,leftmargin=*]
            \item Stance
            \item Stance (Long)
        \end{itemize} & \begin{itemize}[nosep,leftmargin=*]
            \item Outcome complexity (easier)
            \item Outcome complexity (harder)
        \end{itemize} \\
        \addlinespace[0.5em]
        BES Survey Responses & Mellon et al. (2024) & \begin{itemize}[nosep,leftmargin=*]
            \item Categorical: 1 of 6 Most Important Issues (MII)
            \item Categorical: 1 of 12 MII
        \end{itemize} & \begin{itemize}[nosep,leftmargin=*]
            \item MII
            \item MII (Long)
        \end{itemize} & \begin{itemize}[nosep,leftmargin=*]
            \item Outcome complexity (easier)
            \item Outcome complexity (harder)
        \end{itemize} \\
        \addlinespace[0.5em]
        ANES Voter Profiles & Argyle et al. (2023) & \begin{itemize}[nosep,leftmargin=*]
            \item Binary: Biden/Trump (8 Variables)
            \item Binary: Biden/Trump (6 Variables)
        \end{itemize} & \begin{itemize}[nosep,leftmargin=*]
            \item Synthetic
            \item Synthetic (Short)
        \end{itemize} & \begin{itemize}[nosep,leftmargin=*]
            \item Data complexity (easier)
            \item Data complexity (harder)
        \end{itemize} \\
        \bottomrule
    \end{tabular}
    \caption{Summary of datasets, outcomes, and relevant variations used in our approach.}
    \label{tab:datasets}
\end{table}
\end{landscape}

For the US Senator Tweets, we ask the LM to (i) classify the author’s party (Republican vs.\ Democrat) and (ii) detect whether the tweet uses populist rhetoric. We use this pairing to introduce a plausible contrast in task difficulty on the same short-form text: party identification is often signaled by readily available cues (e.g., attacking or criticizing the opposing party), whereas populist language is a more interpretive construct and can be expressed implicitly or in borderline forms. \footnote{We do not claim a universal difficulty ordering. The purpose of this pairing is to induce realistic variation in stability within the same data source, not to assert that partisanship is inherently easier than populism in all settings.}

For the NYT articles, we prompt the LM to classify each article’s overall tone (positive/negative/neutral), first using the full set of articles and then restricting to shorter texts ($\leq$ 1000 words). This contrast probes whether stability changes when the model must compress more versus less information into a single label.

For the UK Manifestos, we ask the LM to code ideology in two formats: a binary left--right label and a 10-point scale. This tests whether stability differs when the same construct is elicited as nominal versus interval output.

For the US Politics Tweets, we compare a binary stance task (dropping ``None'') to a three-category stance task (For/Against/None). The goal is to assess whether stability declines when an explicit neutral/indeterminate category is introduced.

For the BES survey responses, we classify open-ended answers about the ``most important issue'' into either 6 categories or 12 finer-grained categories. This tests how stability changes as the label set becomes more granular and contains closer, potentially overlapping options (e.g., economy-general vs.\ inflation vs.\ living costs).

Finally, for the ANES voter profiles, we compare predictions using a fuller feature set (including explicit partisan and ideological cues) versus a reduced set that omits those political variables. This contrast evaluates how stability changes when key discriminative information is removed from the input.

It is worth noting that for each of these datasets and outcomes, we do not use the exact prompt design as the original article. As such, our tests are not to be understood as an explicit test of the replicability of any of these published findings. 

\subsection{Method}

\subsubsection{Intra-prompt stability}

For each dataset--outcome pairing, we fix a baseline prompt and apply it to the same set of $n$ items across $30$ independent runs.\footnote{Even holding decoding settings fixed, repeated runs can vary when sampling is used; intra-PSS captures this run-to-run variability for a fixed prompt.} Treating each run as a ``rater,'' we compute two complementary intra-prompt stability diagnostics. First, \emph{cumulative} intra-PSS recomputes Krippendorff's $\alpha$ (KA) after each run $j$ using all runs observed up to that point. This yields a stability path that shows both the final pooled agreement level and how quickly the estimate settles as additional runs are added. Second, \emph{adjacent-run} intra-PSS computes KA only on runs $j-1$ and $j$, which provides a local run-to-run diagnostic of volatility for the same prompt. In the examples below, we use up to $n=500$ items per pairing, for up to $15{,}000$ classifications ($500 \times 30$).

We use KA because it is a chance-corrected reliability coefficient that equals $1$ under perfect agreement, approaches $0$ under chance-level agreement, and can be negative when disagreement exceeds chance expectations. For categorical outcomes we use the nominal distance function; for the 10-point ideology outcome we use an interval distance based on squared differences (see Appendix for formal definitions). To estimate uncertainty, we use a nonparametric bootstrap. For cumulative intra-PSS, after each run $j$ we repeatedly resample the annotation records from all runs observed up to $j$, recompute Krippendorff's $\alpha$, and use the resulting bootstrap distribution to form a confidence interval; these intervals therefore track how the cumulative estimate stabilizes as more runs are added. For adjacent-run intra-PSS, we apply the same bootstrap using only runs $j-1$ and $j$, so the interval quantifies uncertainty around the local run-to-run volatility estimate rather than cumulative convergence. For the adjacent-run plots, displayed upper endpoints are bounded at the logical maximum of $1$ (see Appendix).

\subsubsection{Inter-prompt stability}

Inter-prompt stability asks whether an annotation pipeline is robust to small, reasonable changes in prompt wording. We begin from a user-supplied baseline prompt and generate a set of paraphrased variants using an external paraphrasing model, \texttt{PEGASUS} \citep{zhang2019pegasus}. Using an external paraphraser (rather than the annotation model itself) reduces endogeneity and provides a reproducible way to construct prompt variants; the framework does not depend on PEGASUS specifically, and the package also supports manually edited variants.

We control how conservative versus diverse the paraphrases are by varying the paraphraser's decoding \emph{temperature}. At low temperatures, PEGASUS tends to produce close rephrasings that preserve the baseline meaning and resemble plausible alternative wording choices a researcher might make. At higher temperatures, paraphrases become more diverse and may drift in meaning or coherence. We therefore treat low temperatures as the core test of robustness to semantically equivalent rewordings, and higher temperatures as a stress test that intentionally pushes prompts further from the baseline to reveal where the procedure breaks down.

For each paraphraser temperature, we generate multiple prompt variants and apply each variant to the \emph{same} set of items. We then compute Krippendorff's $\alpha$ across the resulting labels, treating the prompt variants as the analogue of ``raters.'' This yields one inter-prompt stability score per temperature. We quantify uncertainty using a nonparametric bootstrap within each temperature by resampling items and recomputing Krippendorff's $\alpha$ (see Appendix).

In the empirical demonstrations below, we generate ten prompt variants per temperature, evaluate twenty-five temperatures from 0.1 to 5.0, and classify up to 500 items per dataset--outcome pairing. A full list of the original prompts and prompt variants that we test is provided in the anonymous Github for this article.\footnote{See: \url{https://github.com/cjbarrie/promptstability_replication}.} We keep the required output format fixed via a constant \texttt{prompt\_postfix} (e.g., binary, categorical, or numeric), so that inter-prompt stability reflects sensitivity to substantive prompt wording rather than differences in response formatting.

In the main results below, we use GPT-3.5 (\texttt{gpt-3.5-turbo}) from OpenAI. For each run of the LM, we keep the temperature of the LM (as opposed to the external paraphraser engine) constant at a conservative 0.1. Importantly, our software package is LM agnostic meaning it is operable across multiple LMs. In the main software package, we include examples of how to implement the main functions using a range of open-source models. In the results below, we also include a limited cross-model comparison between \texttt{gpt-4o} and \texttt{DeepSeek-r1} to illustrate the model-capability dimension in Table~\ref{table:lowicr_ps}; the Appendix provides additional analyses using newer commercial models.

\section{Results}

The results map onto the four sources of instability summarized in Table~\ref{table:lowicr_ps}: criteria ambiguity, outcome complexity, data complexity, and model capability. We begin with an overview of the raw intra- and inter-PSS patterns, then interpret those patterns through the lens of these four dimensions. We treat the filtering exercises below not as a separate theoretical source of instability, but as a practical diagnostic step researchers can use when PSS is low.

\subsection{Overview}

\begin{figure}[htbp]
    \centering
    \includegraphics[width=1\textwidth]{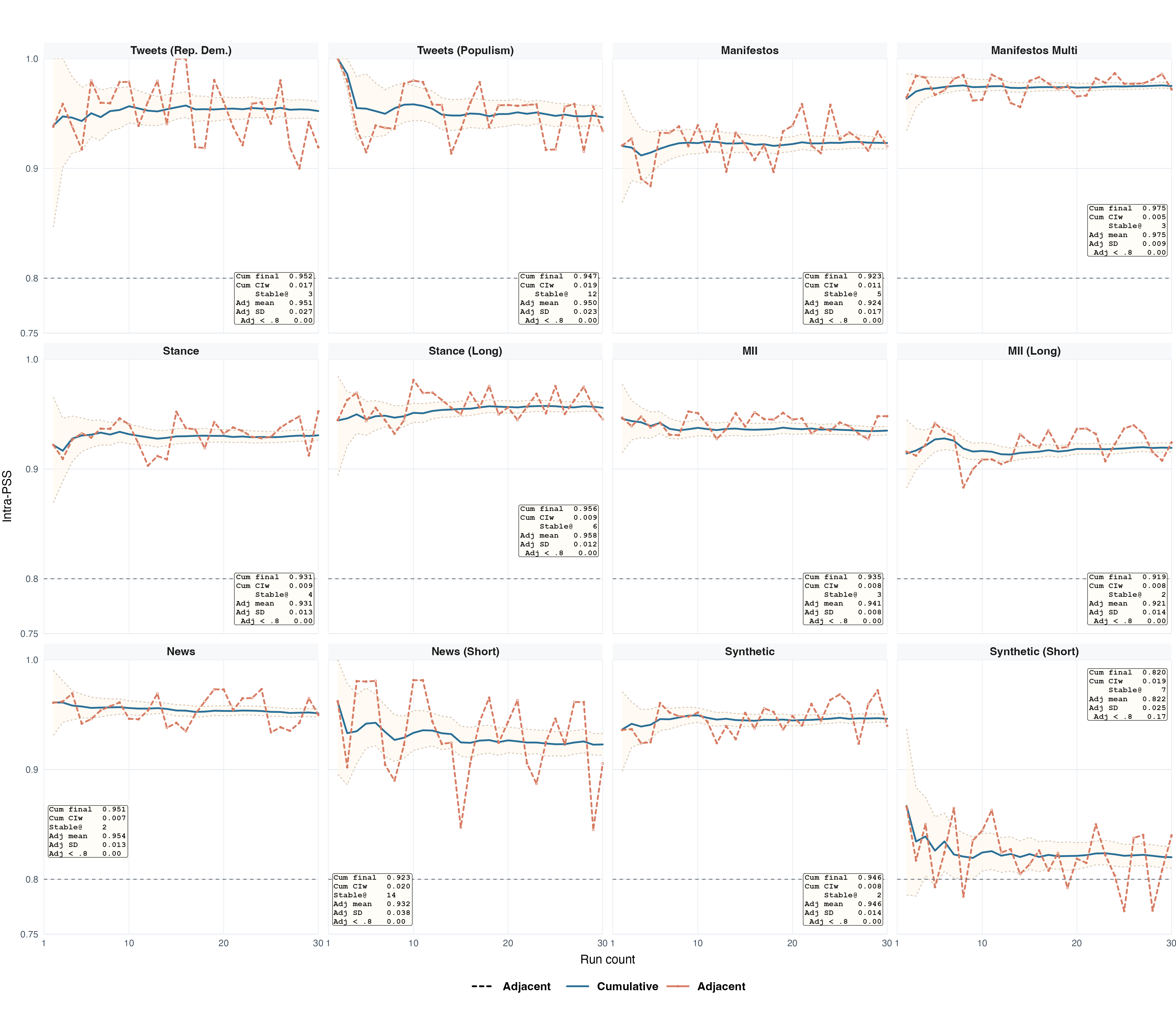}
    \caption{Combined intra-prompt stability scores. In each panel, the solid blue path shows cumulative intra-PSS and the dashed terracotta path shows adjacent-run intra-PSS. Insets report the final cumulative estimate (\texttt{Cum final}), the width of its final bootstrap confidence interval (\texttt{Cum CIw}), the run at which the cumulative point estimate first remains within 0.01 of its final value (\texttt{Stable@}), the mean and standard deviation of adjacent-run intra-PSS across runs (\texttt{Adj mean}, \texttt{Adj SD}), and the share of adjacent-run estimates below 0.8 (\texttt{Adj < .8}). Thus \texttt{Stable@} summarizes estimate convergence rather than CI narrowing. The first two rows show tasks that vary primarily in outcome complexity, while the third row shows tasks that vary primarily in data complexity.}
    \label{fig:combined_within_overlay}
\end{figure}

\begin{figure}[htbp]
    \centering
    \includegraphics[width=1\textwidth]{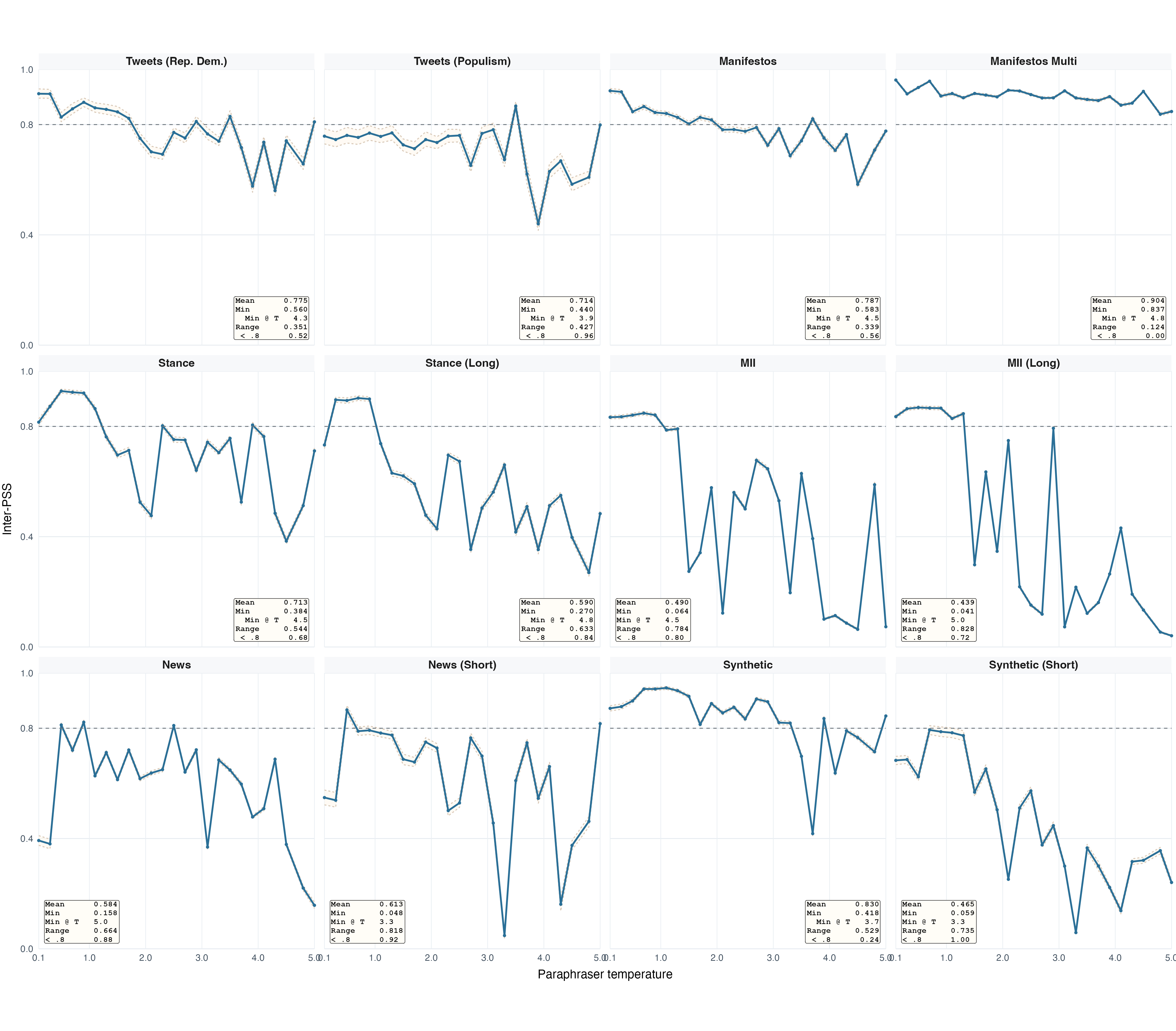}
    \caption{Combined inter-prompt stability scores by paraphraser temperature. Insets report the mean inter-PSS across temperatures (\texttt{Mean}), the minimum observed inter-PSS (\texttt{Min}), the temperature at which that minimum occurs (\texttt{Min @ T}), the range across temperatures (\texttt{Range}), and the share of temperatures below 0.8 (\texttt{< .8}) for each dataset--outcome pairing. The first two rows show tasks that vary primarily in outcome complexity, while the third row shows tasks that vary primarily in data complexity.}
    \label{fig:combined_between_refined}
\end{figure}

We visualize the main results of our tests in Figures \ref{fig:combined_within_overlay} and \ref{fig:combined_between_refined}. In both figures, the first two rows show tasks that vary primarily in outcome complexity, while the third row shows tasks that vary primarily in data complexity. The insets are intended to summarize the two kinds of within-prompt evidence and the temperature-level inter-prompt evidence compactly. In Figure \ref{fig:combined_within_overlay}, each inset reports the final cumulative intra-PSS estimate (\texttt{Cum final}), the width of its final bootstrap confidence interval (\texttt{Cum CIw}), the run at which the cumulative point estimate first remains within 0.01 of its final value (\texttt{Stable@}), the mean and standard deviation of adjacent-run intra-PSS across runs (\texttt{Adj mean}, \texttt{Adj SD}), and the share of adjacent-run estimates below 0.8 (\texttt{Adj < .8}). This means that \texttt{Stable@} is an estimate-convergence statistic: it tells us when the cumulative point estimate has effectively settled near its final level, not when the confidence interval has finished narrowing. In Figure \ref{fig:combined_between_refined}, each inset reports the mean inter-PSS across temperatures (\texttt{Mean}), the minimum observed inter-PSS (\texttt{Min}), the temperature at which that minimum occurs (\texttt{Min @ T}), the range of inter-PSS values across temperatures (\texttt{Range}), and the share of temperatures below 0.8 (\texttt{< .8}).

For the intra-prompt stability analysis---where we repeat the classification task 30 times using the same prompt---the cumulative and adjacent views tell a consistent story.\footnote{Appendix Figure \ref{fig:combined_postpro_within_diagnostics} provides the number of unique annotations across each iteration. A high number of unique annotations indicates that the model has failed to return annotations in the constrained format specified by our prompt. Only the MII dataset suffered materially from this problem.} In most cases, cumulative intra-PSS converges quickly to high agreement, and adjacent-run intra-PSS shows only modest run-to-run fluctuation. The clearest weak point is Synthetic (Short), which has both the lowest final cumulative score and the lowest mean adjacent score. News (Short) and Tweets (Populism) also stand out because their cumulative estimates stabilize more slowly than most other tasks, indicating that repeated runs continue to shift the reliability estimate for longer. At the other end of the distribution, Manifestos Multi is highly stable under both summaries. Taken together, these results suggest that most of the benchmark tasks are repeatable under a fixed prompt, but that low-signal tasks remain vulnerable even before prompts are perturbed.

\begin{comment}
     According to our results, it is clear, first, that practitioners need to pay attention to the format of the LM response. For example, when estimating the intra-PSS on the original annotations, both MII and MII (long) scores are \textit{relatively} a lot lower than their scores when estimating on the filtered data. 
\end{comment}

\subsection{Criteria ambiguity: prompt perturbations}

As for the inter-prompt stability approach, we see in Figure \ref{fig:combined_between_refined} first that the stability of outputs generally decreases as temperature increases. This is as expected: as the prompt becomes more semantically distant from the original, we expect the stability of classifications to reduce as a result of the degradation in quality of coding guidelines (prompts).\footnote{In Appendix Figure \ref{fig:combined_postpro_between_diagnostics}, we display the number of unique annotations by temperature. Generally speaking, the number of unique annotations increases as temperature increases, indicating that the model is more likely to fail to return properly formatted annotations as the prompt quality degrades. }

This pattern speaks most directly to criteria ambiguity. If inter-PSS remains high at low paraphraser temperatures, then the classification routine is robust to the kinds of minor wording changes a researcher might realistically make. If it declines sharply even when prompt variants remain semantically close to the original, then the coding criteria are fragile to small perturbations. That is precisely the pattern we see for several pairings: Stance, Stance (Long), MII, MII (Long), News (Short), News, and Synthetic (Short) all decline rapidly below 0.8 or oscillate substantially at relatively low temperatures.

\subsection{Outcome and data complexity}

The task pairings in Figures \ref{fig:combined_within_overlay} and \ref{fig:combined_between_refined} also help distinguish between instability driven by the outcome and instability driven by the data. In the first two rows, paired tasks vary primarily in outcome complexity while holding the underlying data source fixed. In the third row, paired tasks vary primarily in how much usable signal the input contains.

Several patterns stand out. Stability tends to be highest when both the input is information-rich (lower data complexity) and the target label is sharply defined (lower outcome complexity). It falls earlier---and often more erratically---when either the inputs provide weaker signal or the target construct admits borderline cases. Synthetic (Short) illustrates the first pattern clearly: because each profile contains limited predictive information, stability deteriorates quickly even though the requested output remains binary. The MII and MII (Long) contrast illustrates the second pattern: expanding the label set to more granular and overlapping issue categories makes the classification problem less stable even when the underlying texts are the same type of survey response.

\subsection{Model capability}

\begin{figure}[ht]
    \centering
    \includegraphics[width=1\textwidth]{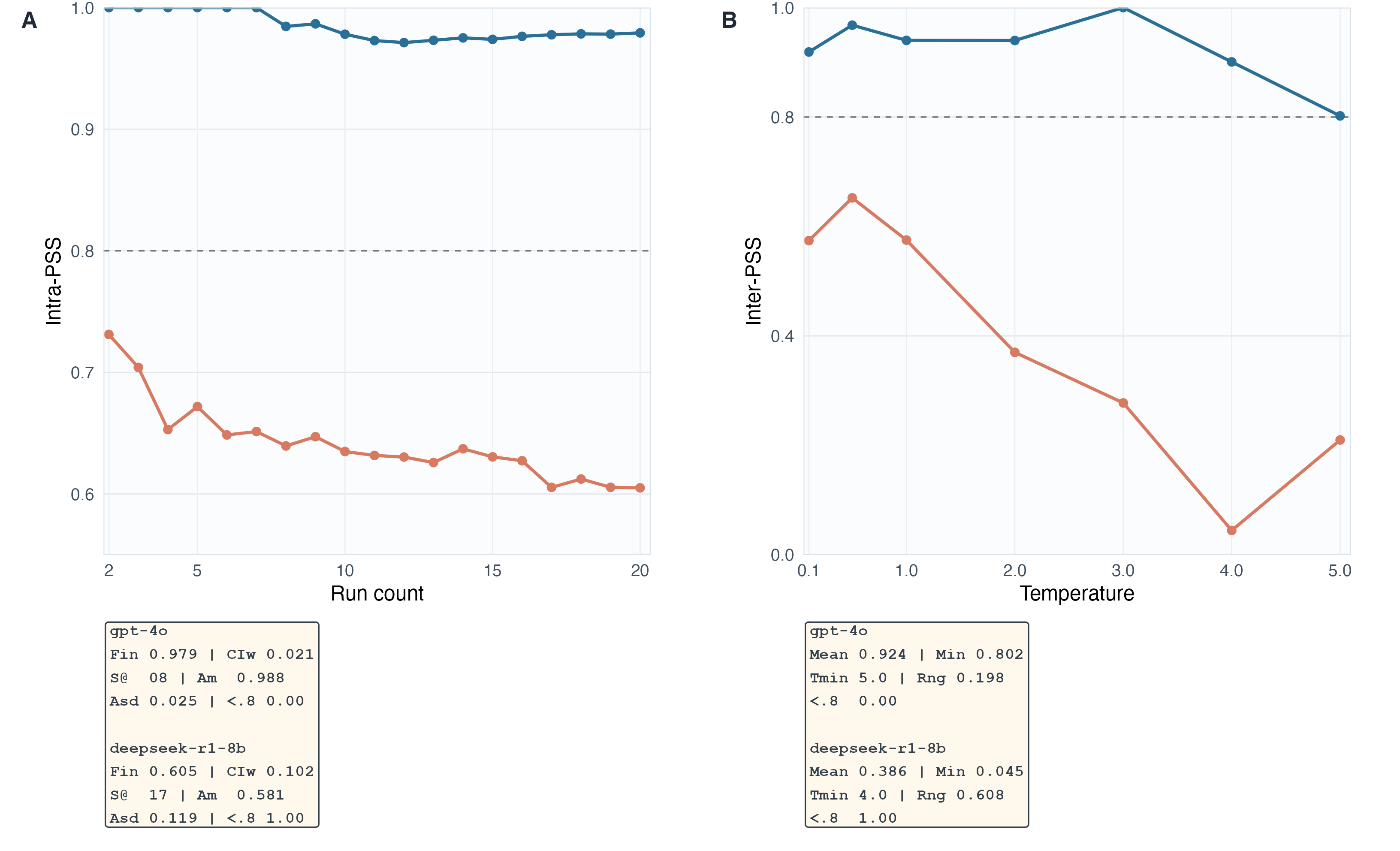}
    \caption{Combined intra- and inter-prompt stability scores for DeepSeek-R1 and GPT-4o.}
    \label{fig:combined_model_comparison}
\end{figure}

The final dimension in Table~\ref{table:lowicr_ps} concerns model capability. To illustrate this directly, Figure \ref{fig:combined_model_comparison} compares a strong closed-source model (\texttt{gpt-4o}) with an open-weight alternative (\texttt{deepseek-r1:8b}) on the Manifestos task. We use this comparison not to claim that one model universally dominates the other, but to show that some instability can plausibly be attributed to model capability rather than only to prompt wording, the data, or the construct.

The comparison suggests that capability differences matter. The intra-PSS for \texttt{deepseek-r1:8b} is lower than for \texttt{gpt-4o}, and its inter-PSS degrades much more quickly as paraphraser temperature increases. In substantive terms, this means that the same task can appear more or less reproducible depending on which LM is used. PSS therefore helps complete the diagnostic framework in Table~\ref{table:lowicr_ps}: low stability may reflect not only ambiguous prompts, difficult data, or complex outcomes, but also limitations of the selected model.

\begin{comment}
    We also see once again the importance of filtering the data. From the Original data, it might appear that LM performance for the MII tasks decreases quickly at low temperatures. However, this is because the LM returned annotations in natural language (e.g., ``This text is about the environment" instead of the integer coding ``40"). 
\end{comment}

We go on to anticipate and answer several questions that applied researchers may have when implementing this approach. We begin with a practical PSS playbook that outlines concrete steps to take when PSS is low versus when it is high, and then turn to a concise diagnostic example.

\subsection{PSS Playbook: Interpreting High vs.\ Low PSS}

In studies of human coder reliability, Krippendorff's $\alpha$ values around $0.8$ are often taken as a useful (though context-dependent) indicator of strong agreement (see \cite{krippendorff2004reliability}). We use the same value as a \emph{heuristic reference point} for prompt stability---not as a hard pass/fail rule. In the within-prompt figure, this means reading \texttt{Cum final} as the overall repeated-run stability level, \texttt{Stable@} as how quickly the cumulative estimate converges, and the adjacent-run summaries as indicators of run-to-run volatility. In the inter-prompt figure, the inset statistics summarize average robustness across paraphraser temperatures (\texttt{Mean}), worst-case sensitivity (\texttt{Min}, \texttt{Min @ T}), spread across temperatures (\texttt{Range}), and how often perturbations push stability below the heuristic benchmark (\texttt{< .8}). The playbook below summarizes typical next steps depending on whether intra- or inter-PSS appears comfortably above, near, or below this benchmark.

\begin{description}[leftmargin=0cm, style=nextline]

    \item[\textbf{Intra-PSS is high (e.g., $\gtrsim 0.8$ early on)}]
    This is consistent with the model behaving repeatably under a fixed prompt and decoding setup.
    In practice, this will usually coincide with a high \texttt{Cum final}, an early \texttt{Stable@}, and adjacent-run summaries that show little run-to-run volatility.
    In most cases, it is reasonable to proceed to the inter-PSS analysis.

    \item[\textbf{Intra-PSS is modest or low (e.g., $\lesssim 0.8$ or slow to stabilize)}]
    Treat this as a warning that repeated runs of the same prompt are not yielding consistent labels.
    In the inset summaries, this will often appear as a lower \texttt{Cum final}, a later \texttt{Stable@}, and/or larger adjacent-run volatility.
    Typical checks include: sharpening decision criteria in the prompt; clarifying the construct (especially edge cases); inspecting whether certain items are inherently ambiguous or noisy; and, if instability persists, considering a more capable model and/or more conservative decoding settings.

    \item[\textbf{Inter-PSS remains high at low paraphraser temperatures (e.g., $\gtrsim 0.8$ up to temperature $\approx 2$)}]
    This pattern is consistent with robustness to minor, semantically close prompt rewordings.
    In the inset, this will usually correspond to a high \texttt{Mean}, a relatively high \texttt{Min}, a modest \texttt{Range}, and a low \texttt{< .8} share.
    At this point, it is usually appropriate to move on to accuracy validation (and any domain-specific substantive checks).

    \item[\textbf{Inter-PSS is low or volatile at low paraphraser temperatures (e.g., $\lesssim 0.8$ up to temperature $\approx 2$)}]
    Treat this as evidence that small prompt perturbations meaningfully change the labeling behavior.
    In the inset, this will often appear as a lower \texttt{Mean}, a low \texttt{Min} reached at relatively mild temperatures, a wider \texttt{Range}, and/or a larger \texttt{< .8} share.
    Common follow-ups are to inspect the specific prompt variants driving instability, revise prompt criteria and output constraints, revisit whether the construct is operationalized tightly enough for the data at hand, and assess whether model choice or decoding settings are contributing.

\end{description}

Overall, higher intra- and inter-PSS provide reassurance that the annotation pipeline is \emph{reproducible} under repeated runs and minor prompt rewordings. Lower PSS does not imply that outputs are inaccurate or biased; rather, it indicates that the procedure is \emph{fragile} to perturbations that are likely to occur in practice, and should be strengthened before investing in resource-intensive validation or downstream inference.

\subsection{Practical diagnostic: output-format filtering}

When PSS is low, a first diagnostic step is to check whether the model is consistently returning the requested output format, since LMs frequently deviate from formatting instructions (see, e.g., \cite{mellon2024ais}). We illustrate this using two dataset--outcome pairings that raise clear red flags in the raw results: MII and Synthetic (Short). For MII, intra-PSS is high but inter-PSS varies substantially even at temperatures below 2; for Synthetic (Short), both intra- and inter-PSS are already concerning.

We therefore apply two simple filters. First, we remove annotations that do not match the required integer format specified in the \texttt{prompt\_postfix}. Second, to make comparisons fair across iterations and temperatures, we construct a filtered and balanced dataset in which each iteration or temperature retains only as many rows as the smallest valid subset after filtering.\footnote{These are not the only options. Researchers might also use regular expressions to parse correct from incorrect outputs, another LM to recode output variants, or structured decoding interfaces such as JSON mode, function calling, grammar-constrained decoding, or logit biasing. These approaches can reduce formatting errors, but they do not eliminate stochasticity in the underlying token probabilities.} The full filtered comparison plots are reported in Appendix Figures~\ref{fig:combined_within_postpro_cumulative_appendix}, \ref{fig:combined_within_postpro_adjacent_appendix}, and~\ref{fig:combined_between_postpro_appendix}.

The filtered results point to two different diagnoses. For MII and MII (Long), both intra- and inter-PSS improve sharply after filtering, and the filtered cumulative and adjacent within trajectories become nearly flat at high agreement (Appendix Figures~\ref{fig:combined_within_postpro_cumulative_appendix}, \ref{fig:combined_within_postpro_adjacent_appendix}, and \ref{fig:combined_between_postpro_appendix}). This strongly suggests that the primary issue is malformed output rather than the prompt, construct, dataset, or model capability. Synthetic (Short) looks very different: filtering and balancing do little to improve stability, which indicates that the deeper problem is the task itself rather than output formatting. In practice, this is exactly how we think filtering should be used: as a fast way to distinguish output-format problems from genuinely unstable classification tasks.

\subsection{How much does this all cost?}

Applied researchers may reasonably object to the approach we outline above that it may be prohibitively expensive. Commercial LM APIs price API calls per 1m input and output tokens.\footnote{A token is a unit used by the model's tokenizer---often a word fragment, but sometimes a whole word, punctuation mark, or whitespace---so token counts vary with the tokenizer and do not correspond exactly to words.} In our empirical tests, we make many tens of thousands of API calls. But how many would an applied researcher need to make? 

\begin{figure}[ht]
    \centering
    \includegraphics[width=1\textwidth]{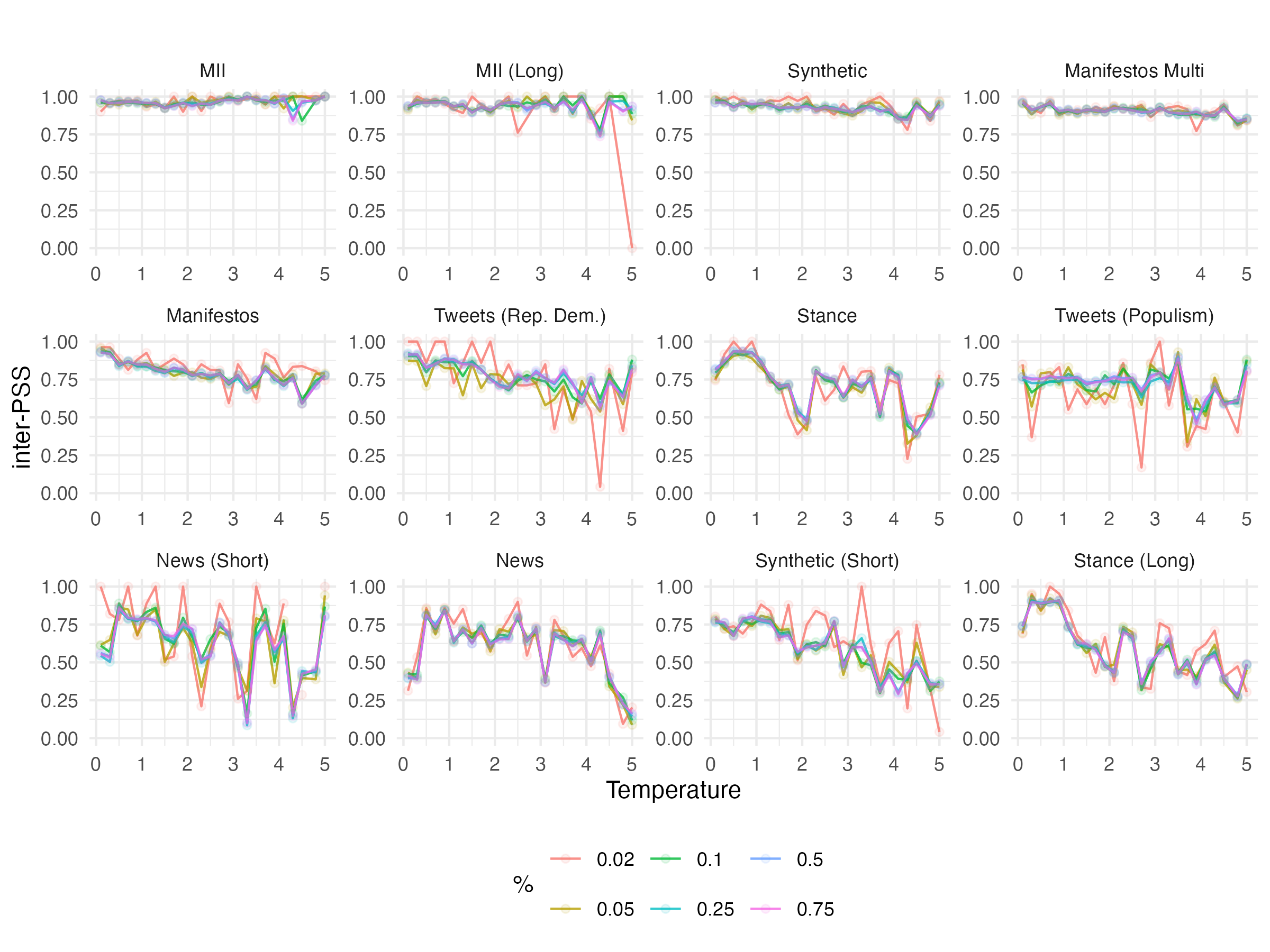}
    \caption{Combined inter-prompt stability scores for random subsamples.}
    \label{fig:combined_between_subsamples}
\end{figure}

To answer this question, we re-estimate the inter-PSS for each dataset but iterate through smaller to larger random subsamples of the (filtered) data. The smallest percentage sample is 2\% and the largest 75\%. We see from Figure \ref{fig:combined_between_subsamples} that even for 2\% of the sample, we are able to recover the basic trends for most datasets.

Appendix Tables \ref{table:dataset_stats} and \ref{table:costs_percentiles} report the full token and cost breakdowns for our analyses, including how costs change when the annotation sample size is reduced. Even with the relatively large number of iterations, temperatures, and prompt variations used here, the total cost for \texttt{gpt-3.5-turbo} remains modest, and a researcher could reproduce a scaled-down version of these checks for roughly \$3 to \$150 depending on the sample size. Appendix Table \ref{table:altlmcosts} reports analogous full-sample cost estimates for a range of open- and closed-source alternatives.

Taken together, these results suggest that the feasibility barrier is modest. In our setting, the full set of analyses using \texttt{gpt-3.5-turbo} costs roughly \$158, while substantially smaller subsamples recover the main stability patterns at much lower cost. We therefore view PSS as feasible as an early-stage diagnostic rather than as a prohibitively expensive validation step. Full token counts, dataset-level cost breakdowns, and alternative model cost estimates are reported in Appendix Tables \ref{table:dataset_stats}, \ref{table:costs_percentiles}, and \ref{table:altlmcosts}.

\subsection{Are the generated prompts actually reliable?}

Our inter-prompt analysis relies on automatically generated prompt variants. We use these variants to \emph{stress-test} robustness, not to certify that a prompt is intrinsically ``good.'' This introduces an obvious concern: some paraphrases, especially at higher temperatures, may drift away from the intended construct. We treat this as an interpretive issue rather than a fatal flaw. If agreement declines only once variants become visibly garbled or semantically off-target, that is consistent with instruction degradation. If it declines even at low temperatures where variants remain coherent and close to the baseline prompt, that is stronger evidence that the annotation pipeline itself is fragile to minor rewording.

Automating prompt variation also reduces researcher degrees of freedom relative to hand-selecting only favorable rephrasings \citep{chang2009reading, grimmer2013text}. At the same time, the package allows manual review and editing when variants are clearly inappropriate. When users do edit variants, we recommend treating this as a documented preprocessing step and reporting the edited prompt list alongside the code. For transparency, the package outputs the full set of prompt variants and their associated stability scores so that researchers can inspect which rephrasings coincide with low agreement.

\section{Conclusion}

This article introduces a diagnostic for LM-based text annotation. We define \emph{prompt stability} as the extent to which a model produces consistent labels when: (i) the same prompt is run repeatedly; and (ii) semantically equivalent prompt variants are used. We operationalize both dimensions with a single agreement statistic treating repeated runs or prompt variants as rater analogues, and we provide an open-source implementation in the \texttt{promptstability} package.

Empirically, we demonstrate the framework across six datasets and twelve outcomes at scale. The resulting stability profiles show that instability can arise even under minor, realistic prompt perturbations. In these settings, high accuracy from a single prompt is not sufficient for reproducibility: if small wording changes or repeated runs shift labels materially, the annotation pipeline is fragile and should be revised before investing in validation.

Several extensions are straightforward. First, future work can more systematically map how decoding choices of the \emph{annotation} model (e.g., temperature and top-$p$) interact with stability. Second, prompt-variant generation could be diversified beyond a single paraphrasing engine, including rule-based perturbations or alternative paraphrasers, to probe robustness to different classes of prompt changes. Third, cross-model comparisons can be expanded to characterize when newer or larger models improve stability and when instability persists because the task is not suitable for an LM annotation pipeline. 

\section{Data Availability Statement}

All code and data used for the above analyses can be found at the anonymized Github repo: \url{https://github.com/cjbarrie/promptstability_replication}. The Python package can be accessed at: \url{https://pypi.org/project/promptstability/}.

\section{AI Use Statement}

AI tools were used in the development of the software package, in code review for statistical analyses, and in redrafting suggestions for the ms. content.

\clearpage
\pagenumbering{arabic}
\setcounter{page}{1}
\renewcommand*{\thepage}{A\arabic{page}}
\appendix
\counterwithin{table}{section}
\counterwithin{figure}{section}

\section*{Appendix} % Add this line to create an unnumbered chapter title

\section{Estimation}

Throughout, we quantify prompt stability using Krippendorff's $\alpha$ (KA). For any set of annotations on the same items, Krippendorff's $\alpha$ is:
\[
\alpha \;=\; 1 - \frac{D_o}{D_e},
\]
where $D_o$ is observed disagreement and $D_e$ is expected disagreement under chance agreement given the marginal label distribution. For nominal (including binary) data, these quantities can be written as:
\[
D_o \;=\; \frac{1}{N} \sum_{c=1}^{n} \sum_{k=1}^{K} \sum_{l=1}^{K} o_{ck}\, o_{cl}\, d_{kl},
\qquad
D_e \;=\; \frac{1}{N(N-1)} \sum_{k=1}^{K} \sum_{l=1}^{K} e_{k}\, e_{l}\, d_{kl},
\]
where $o_{ck}$ is the number of raters assigning category $k$ to item $c$, $e_k$ is the total number of assignments of category $k$ across all items, $d_{kl}$ is the distance between categories $k$ and $l$ (for nominal data, $d_{kl}= \mathbb{I}[k\neq l]$), $K$ is the number of categories, $n$ is the number of items, and $N$ is the total number of rater--item assignments.\footnote{For interval outcomes we use an interval distance (typically squared differences). The implementation uses \texttt{simpledorff}'s nominal or interval metric as appropriate.}

\subsection{Intra-prompt stability (intra-PSS)}

Let $D=\{x_i\}_{i=1}^{n}$ denote the (subsampled) dataset items to be annotated. Fix a single prompt $p$ (optionally $p = \texttt{original\_text} + \texttt{prompt\_postfix}$). The scratch algorithm and the Python implementation treat \emph{repeated runs of the same prompt} as the ``raters.''

Let
\[
C_{i,r}
\]
denote the label produced for item $i$ on run $r$, for $i=1,\ldots,n$ and $r=1,\ldots,R$ (where in the paper we typically set $R=30$; in the package examples this is user-specified via \texttt{iterations}).

After collecting annotations for runs $1,\ldots,r$, the cumulative intra-prompt stability estimate is computed as Krippendorff's $\alpha$ across runs (i.e., across ``raters'') on the same items:
\[
\alpha^{\mathrm{cum}}_r
\;=\;
KA\!\left(
\left\{\, C_{i,1},\, C_{i,2},\, \ldots,\, C_{i,r} \,\right\}_{i=1}^{n}
\right),
\qquad r \ge 2.
\]
In words: for each item $i$, we have $r$ labels from $r$ repeated runs; we compute KA over this item-by-run annotation matrix. This produces the cumulative intra-PSS sequence $\{\alpha^{\mathrm{cum}}_r\}_{r=2}^{R}$.

We also report a complementary adjacent-run diagnostic:
\[
\alpha^{\mathrm{adj}}_r
\;=\;
KA\!\left(
\left\{\, C_{i,r-1},\, C_{i,r} \,\right\}_{i=1}^{n}
\right),
\qquad r \ge 2.
\]
This statistic uses only the labels from runs $r-1$ and $r$, so it captures local run-to-run volatility rather than cumulative convergence. In the revised workflow, the intra-PSS output is therefore the pair of sequences $\{\alpha^{\mathrm{cum}}_r\}_{r=2}^{R}$ and $\{\alpha^{\mathrm{adj}}_r\}_{r=2}^{R}$.

\paragraph{Bootstrapped uncertainty (as implemented).}
To form confidence intervals, the Python code bootstraps by resampling the \emph{annotation records} with replacement from the long-format table of annotations and recomputing KA on each resample. Concretely, let
\[
\mathcal{A}_{\le r} \;=\; \{(i, s, C_{i,s}) : i=1,\ldots,n,\; s=1,\ldots,r\}
\]
denote the set of annotation records up to run $r$ (represented in the code as the subset of rows with \texttt{iteration} $\le r$). For bootstrap draw $b=1,\ldots,B$, the code samples
\[
\mathcal{A}^{(b)}_{\le r} \sim \text{SampleWithReplacement}\bigl(\mathcal{A}_{\le r},\ |\mathcal{A}_{\le r}|\bigr),
\]
computes
\[
\alpha^{(b)}_r \;=\; KA\!\left(\mathcal{A}^{(b)}_{\le r}\right),
\]
and then reports the bootstrap mean and percentile confidence interval:
\[
\widehat{\alpha}_r \;=\; \frac{1}{B}\sum_{b=1}^{B}\alpha^{(b)}_r,
\qquad
\text{CI}_{r,\,0.95} \;=\; \left[\, Q_{0.025}(\{\alpha^{(b)}_r\}),\; Q_{0.975}(\{\alpha^{(b)}_r\}) \,\right].
\]
This matches the package routine \texttt{bootstrap\_krippendorff} used inside \texttt{intra\_pss}. For the adjacent-run statistic, the same bootstrap is applied to the restricted long-format record set
\[
\mathcal{A}^{\mathrm{adj}}_{r} \;=\; \{(i, s, C_{i,s}) : i=1,\ldots,n,\; s\in\{r-1,r\}\},
\]
yielding bootstrap draws
\[
\alpha^{(b),\mathrm{adj}}_r \;=\; KA\!\left(\mathcal{A}^{(b),\mathrm{adj}}_{r}\right).
\]
When we visualize adjacent-run confidence intervals, the reported upper endpoint is bounded above at $1$, the logical maximum of Krippendorff's $\alpha$.

\subsection{Inter-prompt stability (inter-PSS)}

For inter-prompt stability, the scratch algorithm and the Python implementation treat \emph{prompt variants} (generated at a fixed paraphraser temperature) as the ``raters.''

Let $\mathcal{T}$ be the set of paraphraser temperatures. For each $t\in\mathcal{T}$, the code generates $V$ prompt variants
\[
p_{t,1},\, p_{t,2},\, \ldots,\, p_{t,V},
\]
where, in the current implementation, $V = 1 + \texttt{nr\_variations}$ because the list includes the original prompt as one variant plus $nr\_variations$ paraphrases.

For each temperature $t$ and variant $v$, the annotation function is applied to each item $x_i$. Let
\[
C_{i,t,v}
\]
denote the label for item $i$ produced under variant $v$ at paraphraser temperature $t$.

Then inter-prompt stability at temperature $t$ is computed as Krippendorff's $\alpha$ across variants (i.e., across ``raters'') on the same items:
\[
\alpha_t
\;=\;
KA\!\left(
\left\{\, C_{i,t,1},\, C_{i,t,2},\, \ldots,\, C_{i,t,V} \,\right\}_{i=1}^{n}
\right).
\]
The inter-PSS output is the curve $\{(t,\alpha_t)\}_{t\in\mathcal{T}}$.

\paragraph{Multiple annotation repetitions (what the code does).}
The \texttt{inter\_pss} function has an argument \texttt{iterations}. In the common use case (and in our paper figures), we set \texttt{iterations}=1, so each $(i,t,v)$ is annotated once and the notation above is exact.

If \texttt{iterations} $>1$, the current Python implementation appends additional annotation records but does \emph{not} store an explicit repetition index in the output table. KA is then computed on the pooled long-format records with \texttt{annotator\_col} set to \texttt{prompt\_id}. In other words, the estimator remains ``KA across prompt variants within temperature,'' but repeated records for the same $(i,t,v)$ are included as additional rows rather than being aggregated into a single $\widetilde{C}_{i,t,v}$ via a defined rule. (If you want a strict ``one rating per rater per item'' design under repeats, you would need to add an explicit repetition index and an aggregation rule; the present section documents the implementation as-is.)

\paragraph{Bootstrapped uncertainty (as implemented).}
Within each temperature $t$, the code bootstraps by resampling the long-format annotation records for that temperature with replacement, recomputing KA each time. Let
\[
\mathcal{A}_t \;=\; \{(i, v, C_{i,t,v}) : i=1,\ldots,n,\; v=1,\ldots,V\}
\]
denote the set of annotation records at temperature $t$ (represented in the code as \texttt{annotated\_data} for that temperature). For $b=1,\ldots,B$:
\[
\mathcal{A}^{(b)}_{t} \sim \text{SampleWithReplacement}\bigl(\mathcal{A}_t,\ |\mathcal{A}_t|\bigr),
\qquad
\alpha^{(b)}_t = KA\!\left(\mathcal{A}^{(b)}_{t}\right),
\]
and the routine reports the bootstrap mean and percentile confidence interval:
\[
\widehat{\alpha}_t \;=\; \frac{1}{B}\sum_{b=1}^{B}\alpha^{(b)}_t,
\qquad
\text{CI}_{t,\,0.95} \;=\; \left[\, Q_{0.025}(\{\alpha^{(b)}_t\}),\; Q_{0.975}(\{\alpha^{(b)}_t\}) \,\right].
\]
This matches \texttt{bootstrap\_krippendorff} as called inside \texttt{inter\_pss} (with \texttt{annotator\_col} set to \texttt{prompt\_id} and \texttt{experiment\_col} set to \texttt{id}).

\section{Bootstrapping}

We quantify uncertainty in Krippendorff's $\alpha$ using a nonparametric bootstrap. In both the intra- and inter-prompt routines, we work with a long-format annotation table where each row is an annotation record (item id, rater id, and label). For each bootstrap replicate, we resample annotation records with replacement, recompute $\alpha$ on the resampled table, and form percentile confidence intervals from the resulting empirical distribution.

\subsection{Intra-prompt stability}

Let $\mathcal{A}_{\le r}$ denote the set of annotation records obtained after $r$ runs of the fixed prompt (so raters correspond to runs). For each bootstrap replicate $b=1,\ldots,B$, we draw a resample
\[
\mathcal{A}^{(b)}_{\le r} \sim \text{SampleWithReplacement}\!\left(\mathcal{A}_{\le r},\, |\mathcal{A}_{\le r}|\right),
\]
compute
\[
\alpha^{(b)}_{r} = KA\!\left(\mathcal{A}^{(b)}_{\le r}\right),
\]
and report the bootstrap mean and a $95\%$ percentile interval:
\[
\widehat{\alpha}_{r}=\frac{1}{B}\sum_{b=1}^{B}\alpha^{(b)}_{r},
\qquad
\text{CI}_{r,0.95}=\left[Q_{0.025}(\{\alpha^{(b)}_{r}\}),\, Q_{0.975}(\{\alpha^{(b)}_{r}\})\right].
\]
For the adjacent-run diagnostic, we repeat the same procedure on
\[
\mathcal{A}^{\mathrm{adj}}_{r}=\{(i,s,C_{i,s}): i=1,\ldots,n,\; s\in\{r-1,r\}\},
\]
which yields
\[
\widehat{\alpha}^{\mathrm{adj}}_{r}=\frac{1}{B}\sum_{b=1}^{B}\alpha^{(b),\mathrm{adj}}_{r},
\qquad
\text{CI}^{\mathrm{adj}}_{r,0.95}=\left[Q_{0.025}(\{\alpha^{(b),\mathrm{adj}}_{r}\}),\, \min\!\left(Q_{0.975}(\{\alpha^{(b),\mathrm{adj}}_{r}\}),\,1\right)\right].
\]

\subsection{Inter-prompt stability}

Fix a paraphraser temperature $t$ and let $\mathcal{A}_{t}$ denote the set of annotation records produced by applying each of the $V$ prompt variants at temperature $t$ to the same $n$ items (so raters correspond to prompt variants). For each bootstrap replicate $b=1,\ldots,B$, we draw
\[
\mathcal{A}^{(b)}_{t} \sim \text{SampleWithReplacement}\!\left(\mathcal{A}_{t},\, |\mathcal{A}_{t}|\right),
\]
compute
\[
\alpha^{(b)}_{t} = KA\!\left(\mathcal{A}^{(b)}_{t}\right),
\]
and report
\[
\widehat{\alpha}_{t}=\frac{1}{B}\sum_{b=1}^{B}\alpha^{(b)}_{t},
\qquad
\text{CI}_{t,0.95}=\left[Q_{0.025}(\{\alpha^{(b)}_{t}\}),\, Q_{0.975}(\{\alpha^{(b)}_{t}\})\right].
\]

\begin{algorithm}[t]
\caption{Prompt Stability Scoring (PSS): where Krippendorff's $\alpha$ is computed (with bootstrap matching Python implementation)}
\label{alg:pss_ka_where_bootstrap_python}
\begin{algorithmic}[1]
\Require Items $\{x_i\}_{i=1}^n$, annotation function $f(\cdot,\cdot)$, distance metric $d(\cdot,\cdot)$ for KA
\Require (Intra) prompt $p$ and number of runs $J$
\Require (Inter) base prompt $p_0$, temperatures $\mathcal{T}$, variants per temperature $V$
\Require Bootstrap samples $B$ (optional)
\vspace{2mm}

\Statex \textbf{Intra-prompt stability (intra-PSS)}
\State Initialize long-format annotation table $\mathcal{A} \gets \emptyset$
\Comment{Each row is a record $(i,\texttt{run}=j, y)$}
\For{$j = 1$ \textbf{to} $J$}
    \For{$i = 1$ \textbf{to} $n$}
        \State $y \gets f(x_i, p)$
        \State Append record $(i, j, y)$ to $\mathcal{A}$
    \EndFor
    \If{$j \ge 2$}
        \State \textbf{Compute Krippendorff's $\alpha$ across runs (raters)}:
        \State \hspace{5mm} $\alpha_j \gets KA\!\left(\mathcal{A}_{\le j};\, d\right)$
        \Comment{$\mathcal{A}_{\le j}$ are records with $\texttt{run}\le j$}
        \If{$B > 0$}
            \For{$b = 1$ \textbf{to} $B$}
                \State Sample annotation records with replacement:
                \State \hspace{5mm} $\mathcal{A}^{(b)}_{\le j} \sim \text{SampleWithReplacement}\!\left(\mathcal{A}_{\le j},\,|\mathcal{A}_{\le j}|\right)$
                \State $\alpha^{(b)}_j \gets KA\!\left(\mathcal{A}^{(b)}_{\le j};\, d\right)$
            \EndFor
            \State Form percentile CI for $\alpha_j$ from $\{\alpha^{(b)}_j\}_{b=1}^B$
            \State (Optionally) report bootstrap mean $\widehat{\alpha}_j = \frac{1}{B}\sum_{b=1}^B \alpha^{(b)}_j$
        \EndIf
    \EndIf
\EndFor
\Statex \Comment{Output: sequence $\{\alpha_j\}_{j=2}^J$ (and CIs if bootstrapped)}

\vspace{3mm}
\Statex \textbf{Inter-prompt stability (inter-PSS)}
\For{\textbf{each} paraphraser temperature $t \in \mathcal{T}$}
    \State Generate prompt variants $\{p_{t,v}\}_{v=1}^{V}$ (optionally include $p_0$ as one variant)
    \State Initialize long-format annotation table $\mathcal{A}_t \gets \emptyset$
    \Comment{Each row is a record $(i,\texttt{variant}=v, y)$ at temperature $t$}
    \For{$v = 1$ \textbf{to} $V$}
        \For{$i = 1$ \textbf{to} $n$}
            \State $y \gets f(x_i, p_{t,v})$
            \State Append record $(i, v, y)$ to $\mathcal{A}_t$
        \EndFor
    \EndFor
    \State \textbf{Compute Krippendorff's $\alpha$ across prompt variants (raters), within temperature $t$}:
    \State \hspace{5mm} $\alpha_t \gets KA\!\left(\mathcal{A}_t;\, d\right)$
    \If{$B > 0$}
        \For{$b = 1$ \textbf{to} $B$}
            \State Sample annotation records with replacement:
            \State \hspace{5mm} $\mathcal{A}^{(b)}_{t} \sim \text{SampleWithReplacement}\!\left(\mathcal{A}_t,\,|\mathcal{A}_t|\right)$
            \State $\alpha^{(b)}_t \gets KA\!\left(\mathcal{A}^{(b)}_{t};\, d\right)$
        \EndFor
        \State Form percentile CI for $\alpha_t$ from $\{\alpha^{(b)}_t\}_{b=1}^B$
        \State (Optionally) report bootstrap mean $\widehat{\alpha}_t = \frac{1}{B}\sum_{b=1}^B \alpha^{(b)}_t$
    \EndIf
\EndFor
\Statex \Comment{Output: curve $\{(t,\alpha_t)\}_{t\in\mathcal{T}}$ (and CIs if bootstrapped)}
\end{algorithmic}
\end{algorithm}

\section{Additional figures}

\begin{landscape}
\begin{figure}[ht]
    \centering
    \includegraphics[width=1.2\textwidth]{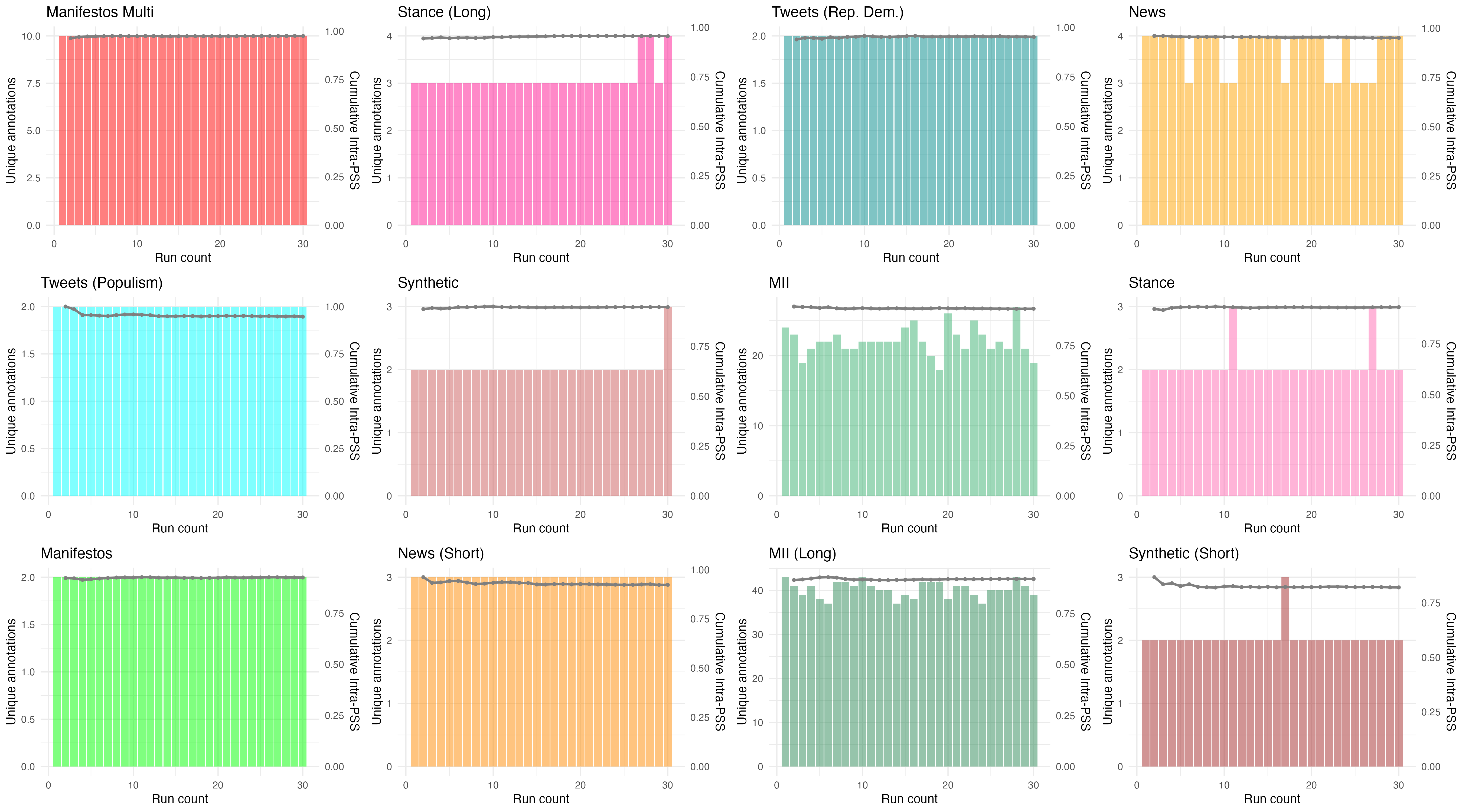}
    \caption{Unique annotations by dataset and iteration.}
    \label{fig:combined_postpro_within_diagnostics}
\end{figure}

\begin{figure}[ht]
    \centering
    \includegraphics[width=1.2\textwidth]{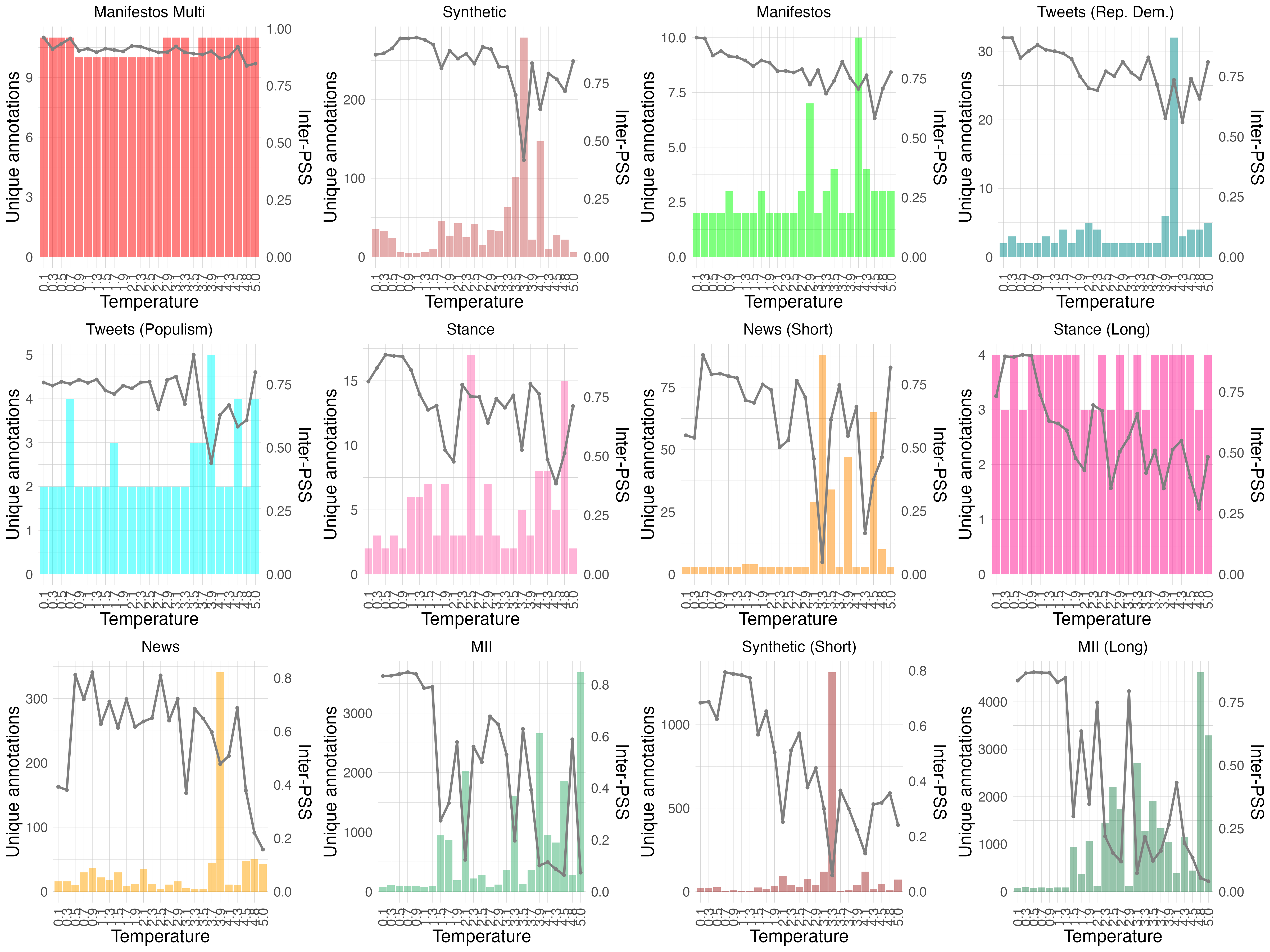}
    \caption{Unique annotations by dataset and temperature.}
    \label{fig:combined_postpro_between_diagnostics}
\end{figure}

\end{landscape}

\section{LM Costs}

\begin{landscape}
  \begin{table}[!h]
    \centering
    \resizebox{\ifdim\width>\linewidth\linewidth\else\width\fi}{!}{%
      \begin{tabular}{lrrrr}
        \toprule
        Dataset           & Total Rows & Input tokens total & Output tokens total & Avg.\ tokens/call \\
        \midrule
        Manifestos        & 276{,}347    & 25{,}968{,}922           & 276{,}347              & 94               \\
        Manifestos Multi  & 276{,}201    & 25{,}954{,}819           & 276{,}201              & 94               \\
        MII               & 251{,}378    & 458{,}994             & 251{,}378              & 2                \\
        MII (Long)        & 233{,}670    & 462{,}566             & 233{,}670              & 2                \\
        News              & 409{,}734    & 198{,}099{,}296          & 409{,}734              & 483              \\
        News (Short)      & 84{,}451     & 10{,}257{,}657           & 84{,}451               & 121              \\
        Stance            & 412{,}362    & 6{,}881{,}473            & 412{,}362              & 17               \\
        Stance (Long)     & 264{,}775    & 4{,}459{,}825            & 264{,}775              & 17               \\
        Synthetic         & 395{,}563    & 16{,}240{,}554           & 395{,}563              & 41               \\
        Synthetic (Short) & 374{,}733    & 11{,}604{,}411           & 374{,}733              & 31               \\
        Tweets (Populism) & 82{,}405     & 3{,}001{,}132            & 82{,}405               & 36               \\
        Tweets (Rep. Dem.)& 82{,}403     & 3{,}001{,}306            & 82{,}403               & 36               \\
        Total             & 3{,}144{,}022   & 306{,}390{,}955          & 3{,}144{,}022             & ---              \\
        \bottomrule
      \end{tabular}%
    }
    \caption{Total size and tokens by dataset}
    \label{table:dataset_stats}
  \end{table}
  
  \vspace{2em}
  
  \begin{table}[!h]
    \centering
    \resizebox{\ifdim\width>\linewidth\linewidth\else\width\fi}{!}{%
      \begin{tabular}{lrrrrrrrrr}
        \toprule
        Dataset           & Input cost & Output cost & Total cost & 2\%  & 5\%   & 10\%  & 25\%  & 50\%  & 75\% \\
        \midrule
        Manifestos        & \$12.98      & \$0.41        & \$13.40      & 5{,}527 & 13{,}817 & 27{,}635 & 69{,}087 & 138{,}174 & 207{,}260 \\
        Manifestos Multi  & \$12.98      & \$0.41        & \$13.39      & 5{,}524 & 13{,}810 & 27{,}620 & 69{,}050 & 138{,}100 & 207{,}151 \\
        MII               & \$0.23       & \$0.38        & \$0.61       & 5{,}028 & 12{,}569 & 25{,}138 & 62844 & 125{,}689 & 188{,}534 \\
        MII (Long)        & \$0.23       & \$0.35        & \$0.58       & 4{,}673 & 11{,}684 & 23{,}367 & 58418 & 116{,}835 & 175{,}252 \\
        News              & \$99.05      & \$0.61        & \$99.66      & 8{,}195 & 20{,}487 & 40{,}973 & 102{,}434 & 204{,}867 & 307{,}300 \\
        News (Short)      & \$5.13       & \$0.13        & \$5.26       & 1{,}689 & 4{,}223  & 8{,}445  & 21{,}113  & 42{,}226  & 63{,}338  \\
        Stance            & \$3.44       & \$0.62        & \$4.06       & 8{,}247 & 20{,}618 & 41{,}236 & 103{,}090 & 206{,}181 & 309{,}272 \\
        Stance (Long)     & \$2.23       & \$0.40        & \$2.63       & 5{,}296 & 13{,}239 & 26{,}478 & 66{,}194  & 132{,}388 & 198{,}581 \\
        Synthetic         & \$8.12       & \$0.59        & \$8.71       & 7{,}911 & 19{,}778 & 39{,}556 & 98{,}891  & 197{,}782 & 296{,}672 \\
        Synthetic (Short) & \$5.80       & \$0.56        & \$6.36       & 7{,}495 & 18{,}737 & 37{,}473 & 93{,}683  & 187{,}366 & 281{,}050 \\
        Tweets (Populism) & \$1.50       & \$0.12        & \$1.62       & 1{,}648 & 4{,}120  & 8{,}240  & 20{,}601  & 41{,}202  & 61{,}804  \\
        Tweets (Rep. Dem.)& \$1.50       & \$0.12        & \$1.62       & 1{,}648 & 4{,}120  & 8{,}240  & 20{,}601  & 41{,}202  & 61{,}802  \\
        Total             & \$153.20     & \$4.72        & \$157.91     & \$3.16 & \$7.90  & \$15.79 & \$39.48  & \$78.96  & \$118.43 \\
        \bottomrule
      \end{tabular}%
    }
    \caption{Cost estimates for each dataset by percentile. Note these costs are current as of August, 2025 but provide a general rough estimate.}
    \label{table:costs_percentiles}
  \end{table}
\end{landscape}

\begin{table}[!h]
\centering
\caption{Estimated Total API Cost for Various Models}
\centering
\resizebox{\ifdim\width>\linewidth\linewidth\else\width\fi}{!}{
\begin{tabular}[t]{lrrlll}
\toprule
Model & input\_price & output\_price & cost\_input & cost\_output & total\_cost\\
\midrule
gpt-4o-mini & 0.15 & 0.60 & 45.96 & 1.89 & 47.85\\
gpt-4o & 2.50 & 10.00 & 765.98 & 31.44 & 797.42\\
o1-mini & 1.10 & 4.40 & 337.03 & 13.83 & 350.86\\
o1 & 15.00 & 60.00 & 4595.86 & 188.64 & 4784.51\\
o3-mini & 1.10 & 4.40 & 337.03 & 13.83 & 350.86\\
\addlinespace
claude-3.5-haiku & 0.80 & 4.00 & 245.11 & 12.58 & 257.69\\
claude-3.5-sonnet & 3.00 & 15.00 & 919.17 & 47.16 & 966.33\\
deepseek-v3 & 0.14 & 0.28 & 42.89 & 0.88 & 43.78\\
deepseek-r1 & 0.55 & 2.19 & 168.52 & 6.89 & 175.4\\
gemini-1.5-flash & 0.15 & 0.60 & 45.96 & 1.89 & 47.85\\
\addlinespace
gemini-1.5-pro & 1.25 & 5.00 & 382.99 & 15.72 & 398.71\\
mistral-small & 0.20 & 0.60 & 61.28 & 1.89 & 63.16\\
mistral-large & 2.00 & 6.00 & 612.78 & 18.86 & 631.65\\
\bottomrule
\end{tabular}}
\label{table:altlmcosts}
\end{table}

\section{Additional analyses}

\subsection{Filtered diagnostic comparisons}

In the main text, we treat filtering as a practical diagnostic rather than as a separate theoretical source of instability. Here we report the full filtered comparisons. We apply two procedures. First, we remove annotations that do not match the required integer format specified in the \texttt{prompt\_postfix}. Second, we construct a filtered and balanced dataset in which each iteration or temperature retains only as many rows as the smallest valid subset after filtering. This allows us to compare the original results to versions in which malformed outputs have been removed, with and without balancing the sample across iterations or temperatures.

The main takeaway is that these comparisons are diagnostic rather than substantive in themselves. For MII, filtering substantially improves both intra- and inter-PSS, suggesting that much of the apparent instability is driven by output-format errors. For Synthetic (Short), by contrast, filtering has little effect, indicating that the underlying task remains unstable even after malformed outputs are removed.

\begin{figure}[ht]
    \centering
    \includegraphics[width=1\textwidth]{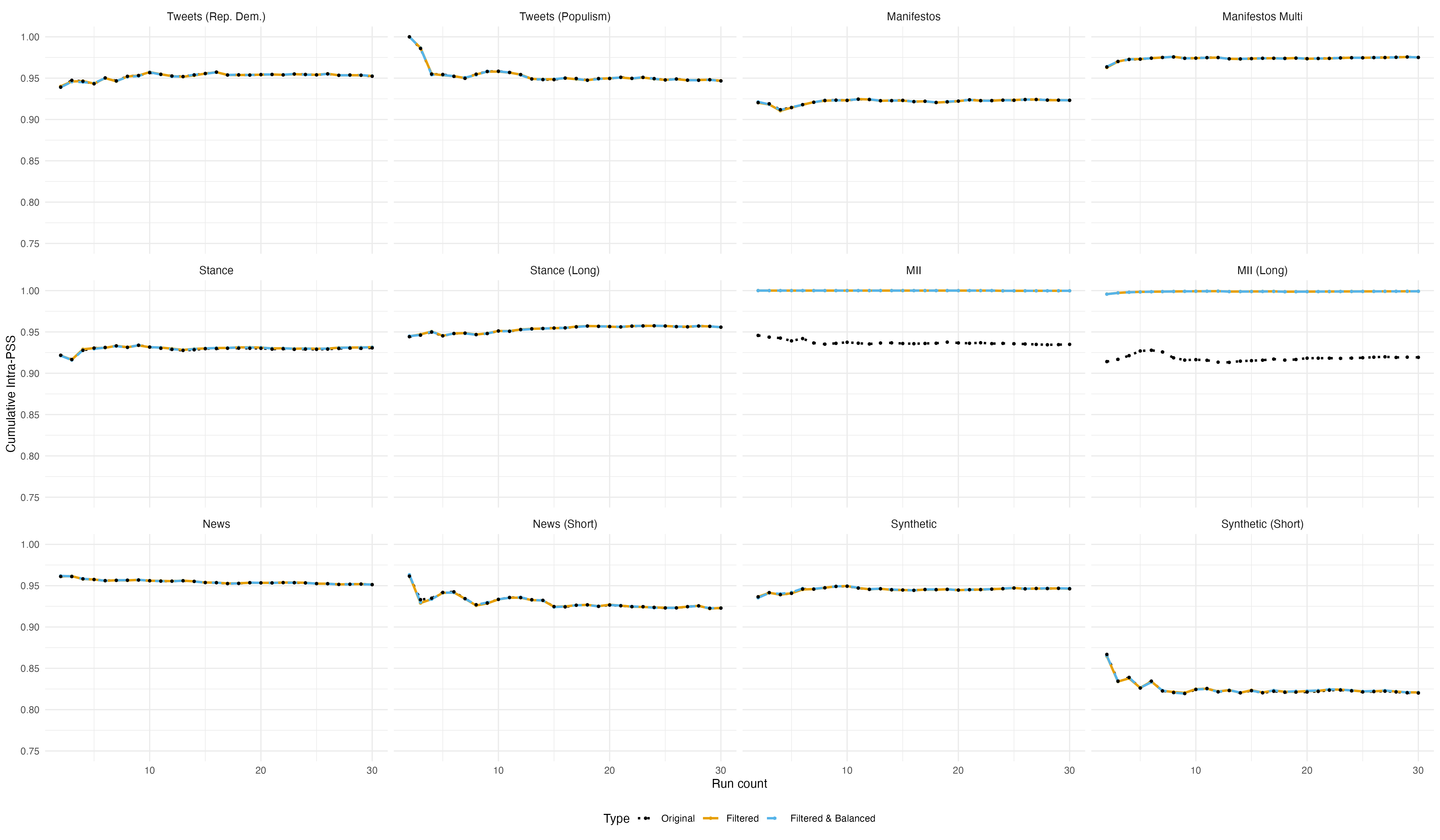}
    \caption{Filtered diagnostic comparison for cumulative intra-prompt stability.}
    \label{fig:combined_within_postpro_cumulative_appendix}
\end{figure}

\begin{figure}[ht]
    \centering
    \includegraphics[width=1\textwidth]{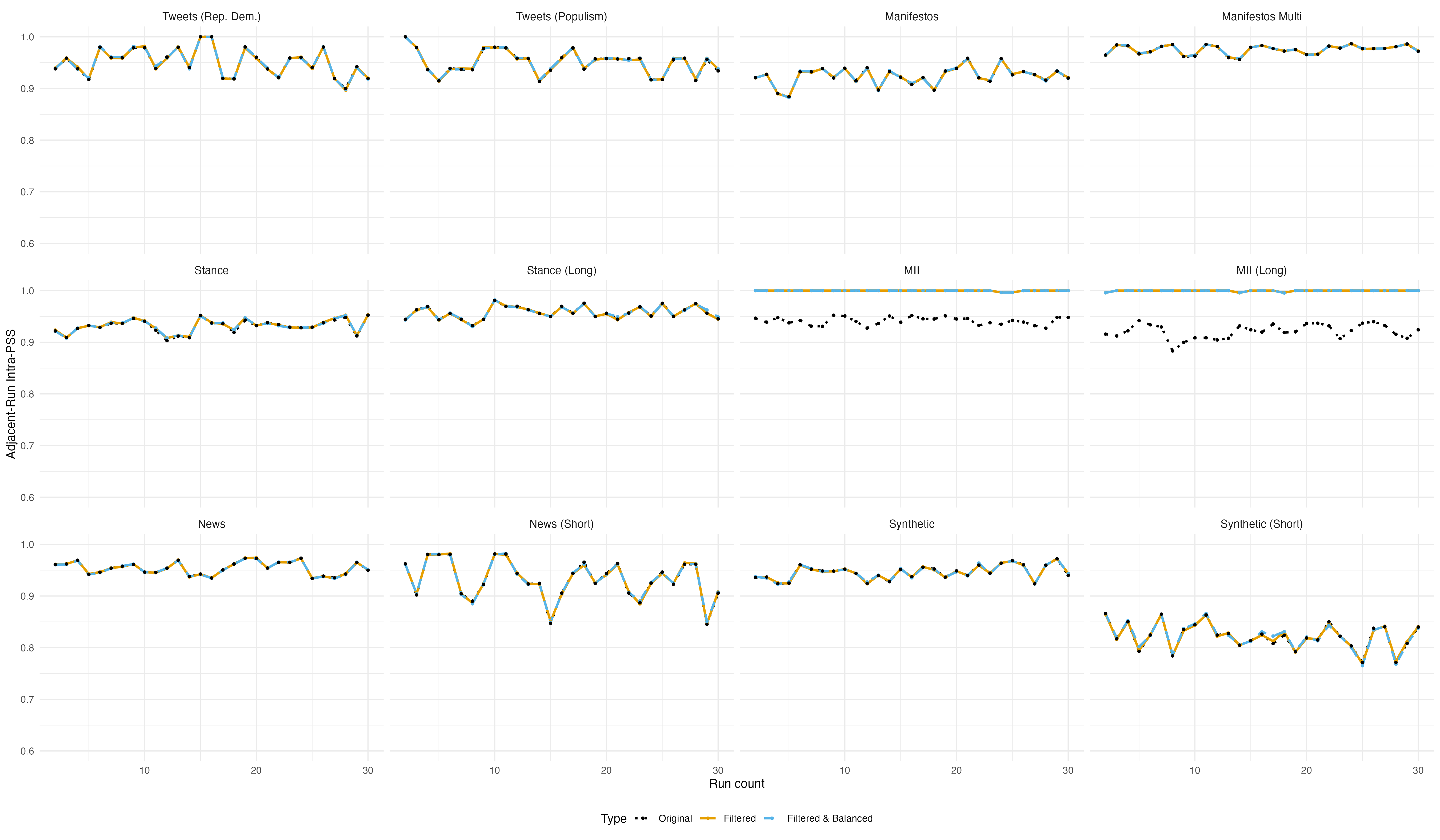}
    \caption{Filtered diagnostic comparison for adjacent-run intra-prompt stability.}
    \label{fig:combined_within_postpro_adjacent_appendix}
\end{figure}

\begin{figure}[ht]
    \centering
    \includegraphics[width=1\textwidth]{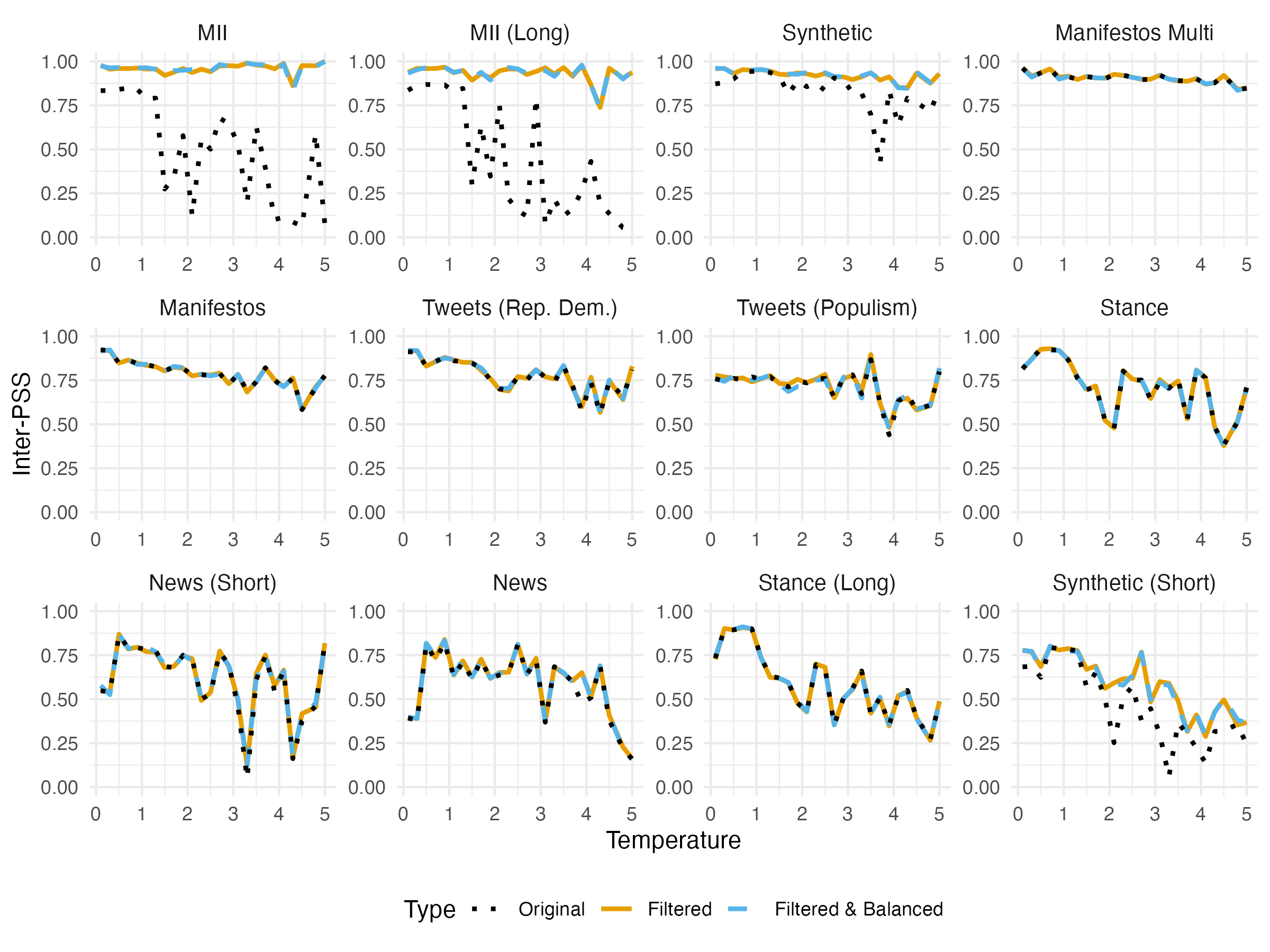}
    \caption{Filtered diagnostic comparison for inter-prompt stability.}
    \label{fig:combined_between_postpro_appendix}
\end{figure}

\subsection{Are other LMs more stable?}

\begin{figure}[ht]
    \centering
    \includegraphics[width=1\textwidth]{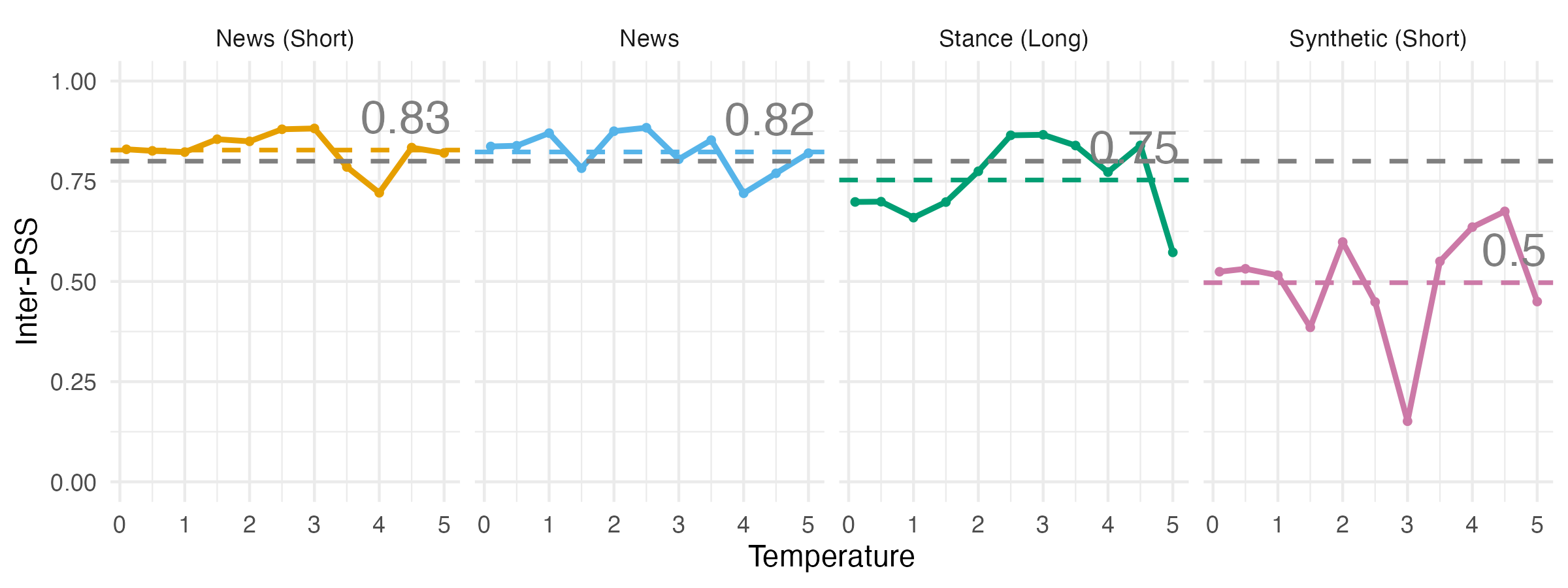}
    \caption{Combined inter-prompt stability scores for lowest inter-PSS outcomes in post-filtering analyses.}
    \label{fig:combined_between_updated}
\end{figure}

In our main analyses, we find that---for some datasets---the inter-PSS decreases even at low temperatures. One might expect that the latest commercial LMs would perform better than the LM we use (\texttt{gpt-3.5-turbo}), which has since been surpassed by newer models. To test this conjecture, we take the four worst (post-filtering) performing dataset and outcome pairings (News (Short); News; Stance (Long); and Synthetic (Short) and re-estimate the inter-PSS using \texttt{gpt-4o-mini}. We see in Figure \ref{fig:combined_between_updated} that while this model produces annotations that are indeed more stable, the Synthetic (Short) data is about as stable as in the main analyses. 

It should be noted that while PSS is not a measure of accuracy, it provides a crucial dimension of model evaluation: the stability and reproducibility of a model's outputs. When choosing between language models, researchers can compare their intra- and inter-PSS profiles to assess which model generates more stable classification for a given task. A model that produces highly unstable outputs cannot be trusted to yield accurate classifications, regardless of its average performance on benchmark datasets. We therefore recommend that researchers treat PSS as one dimension of model selection, alongside accuracy, cost, speed, and domain suitability. The main text provides a limited cross-model comparison between \texttt{gpt-4o} and \texttt{deepseek-r1:8b}; the appendix analysis here is intended as a supplementary robustness check using a newer commercial model.

\section{Prompt variants}

The prompt variants we used for the main analyses can be found at the anonymized Github repo: \url{https://github.com/cjbarrie/promptstability_replication}.

\tiny
\clearpage
\section{Original prompts}

\begin{lstlisting}[backgroundcolor=\color{white},   
    basicstyle=\ttfamily\footnotesize,
    breaklines=true, 
    breakatwhitespace=true            
    captionpos=b,                    
    keepspaces=true,                 
    language=bash, 
    stringstyle=\color{black},
    caption={Original prompts},
    label=lst:originalprompts]
% Format: Dataset 	 Prompt
tweets_rd_within.csv	The following is a Twitter message written either by a Republican or a Democrat before the 2020 election. Your task is to guess whether the author is Republican or Democrat. [Respond 0 for Democrat, or 1 for Republican. Guess if you do not know. Respond nothing else.]
tweets_pop_within.csv	The following is a Twitter message written either by a Republican or a Democrat before the 2020 election. Your task is to label whether or not it contains populist language. [Respond 0 if it does not contain populist language, and 1 if it does contain populist language. Guess if you do not know. Respond nothing else.]
news_within.csv	The text provided is some newspaper text. Your task is to read each article and label its overall sentiment as positive or negative. Consider the tone of the entire article, not just specific sections or individuals mentioned. [Respond 0 for negative, 1 for positive, and 2 for neutral. Respond nothing else.]
news_short_within.csv	The text provided is some newspaper text. Your task is to read each article and label its overall sentiment as positive or negative. Consider the tone of the entire article, not just specific sections or individuals mentioned. [Respond 0 for negative, 1 for positive, and 2 for neutral. Respond nothing else.]
manifestos_within.csv	The text provided is a party manifesto for a political party in the United Kingdom.     Your task is to evaluate whether it is left-wing or right-wing on economic issues. Respond with 0 for left-wing, or 1 for right-wing.     Respond nothing else.
manifestos_multi_within.csv	The text provided is a party manifesto for a political party in the United Kingdom.     Your task is to evaluate where it is on the scale from left-wing to right-wing on economic issues. Respond with a number from 1 to 10. 1 corresponds to most left-wing. 10 corresponds to most right-wing.     Respond nothing else.
stance_long_within.csv	The text provided come from some tweets about Donald Trump. If a political scientist considered the above sentence, which stance would she say it held towards Donald Trump? [Respond 0 for negative, and 1 for positive. Respond nothing else.]
stance_long_within.csv	The text provided come from some tweets about Donald Trump. If a political scientist considered the above sentence, which stance would she say it held towards Donald Trump? [Respond 0 for negative, 1 for positive, and 2 for none. Respond nothing else.]
mii_within.csv	Here are some open-ended responses from a scientific study of voters to the question what is the most important issue facing the country?. Please assign one of the following categories to each open ended text response. [Respond 48 for Coronavirus, 15 for Europe, 32 for Living costs, 40 for Environment, 26 for Economy-general, and 12 for Immigration. Respond nothing else.]
mii_long_within.csv	Here are some open-ended responses from a scientific study of voters to the question what is the most important issue facing the country?. Please assign one of the following categories to each open ended text response. [Respond 48 for Coronavirus, 15 for Europe, 32 for Living costs, 40 for Environment, 26 for Economy-general, 12 for Immigration, 4 for Pol-neg i.e., complaints about politics, 1 for Health, 31 for Inflation, 22 for War, 5 for Partisan-neg i.e., complaints about a party of politician, and 14 for Crime. Respond nothing else.]
synth_within.csv	In the 2020 presidential election, Donald Trump is the Republican candidate, and Joe Biden is the Democratic candidate. The following is some information about an individual voter. I want you to tell me how you think they voted. [Respond 0 for Biden, or 1 for Trump. Guess if you do not know. Respond nothing else.]
synth_short_within.csv	In the 2020 presidential election, Donald Trump is the Republican candidate, and Joe Biden is the Democratic candidate. The following is some information about an individual voter. I want you to tell me how you think they voted. [Respond 0 for Biden, or 1 for Trump. Guess if you do not know. Respond nothing else.]
\end{lstlisting}

\end{document}